\documentclass[10pt,journal,compsoc]{IEEEtran}

\ifCLASSOPTIONcompsoc
  \usepackage[nocompress]{cite}
\else
  \usepackage{cite}
\fi

\ifCLASSINFOpdf

\else

\fi

\usepackage{amsmath,amsfonts}
\usepackage{algorithmic}
\usepackage{algorithm}
\usepackage{array}
\usepackage{mathtools}
\usepackage[caption=false,font=normalsize,labelfont=sf,textfont=sf]{subfig}
\usepackage{textcomp}
\usepackage{stfloats}
\usepackage{url}
\usepackage{verbatim}
\usepackage{graphicx}
\usepackage{cite}
\usepackage{times}
\usepackage{epsfig}
\usepackage{graphicx}
\usepackage{amsmath}
\usepackage{amssymb}
\usepackage{booktabs}
\usepackage{makecell}
\usepackage{arydshln}
\usepackage{enumitem}
\usepackage[table]{xcolor}
\usepackage{pifont}
\newcommand{\cmark}{\ding{51}}%
\usepackage{multirow}
\usepackage{caption}

\newcommand{\myparagraph}[1]{\vspace{3pt}\noindent\textbf{#1}}
\newcommand\our{\text{Mask2Anomaly}}

\usepackage[pagebackref=true,breaklinks=true,letterpaper=true,colorlinks,bookmarks=false]{hyperref}
\usepackage[capitalize]{cleveref}
\crefname{section}{Sec.}{Secs.}
\Crefname{section}{Section}{Sections}
\Crefname{table}{Table}{Tables}
\crefname{table}{Tab.}{Tabs.}
\begin{document}

\title{Mask2Anomaly: Mask Transformer for Universal Open-set Segmentation}

\author{Shyam Nandan Rai,
        Fabio Cermelli,
        Barbara Caputo,
        and Carlo Masone
\IEEEcompsocitemizethanks{\IEEEcompsocthanksitem S. N. Rai, F. Cermelli, B. Caputo, and C. Masone are with Politecnico di Torino.
E-mail: \{name.surname\}@polito.it
}

}



\IEEEtitleabstractindextext{
\begin{abstract}
Segmenting unknown or anomalous object instances is a critical task in autonomous driving applications, and it is approached traditionally as a per-pixel classification problem. However, reasoning individually about each pixel without considering their contextual semantics results in high uncertainty around the objects' boundaries and numerous false positives. We propose a paradigm change by shifting from a per-pixel classification to a mask classification. Our mask-based method, {\our}, demonstrates the feasibility of integrating a mask-classification architecture to jointly address anomaly segmentation, open-set semantic segmentation, and open-set panoptic segmentation. {\our} includes several technical novelties that are designed to improve the detection of anomalies/unknown objects: i) a global masked attention module to focus individually on the foreground and background regions; ii) a mask contrastive learning that maximizes the margin between an anomaly and known classes; iii) a mask refinement solution to reduce false positives; and iv) a novel approach to mine unknown instances based on the mask-
architecture properties. By comprehensive qualitative and qualitative evaluation, we show Mask2Anomaly achieves new state-of-the-art results across the benchmarks of anomaly segmentation, open-set semantic segmentation, and open-set panoptic segmentation. 
\end{abstract}
\begin{IEEEkeywords}
Anomaly Segmentation, Open-set Semantic Segmentation, Open-set Panoptic Segmentation, Mask Transformers.
\end{IEEEkeywords}}
\maketitle

\section{Introduction}
\label{sec:intro}
\begin{figure}[t]
    \begin{center}
        \includegraphics[width=1\linewidth]{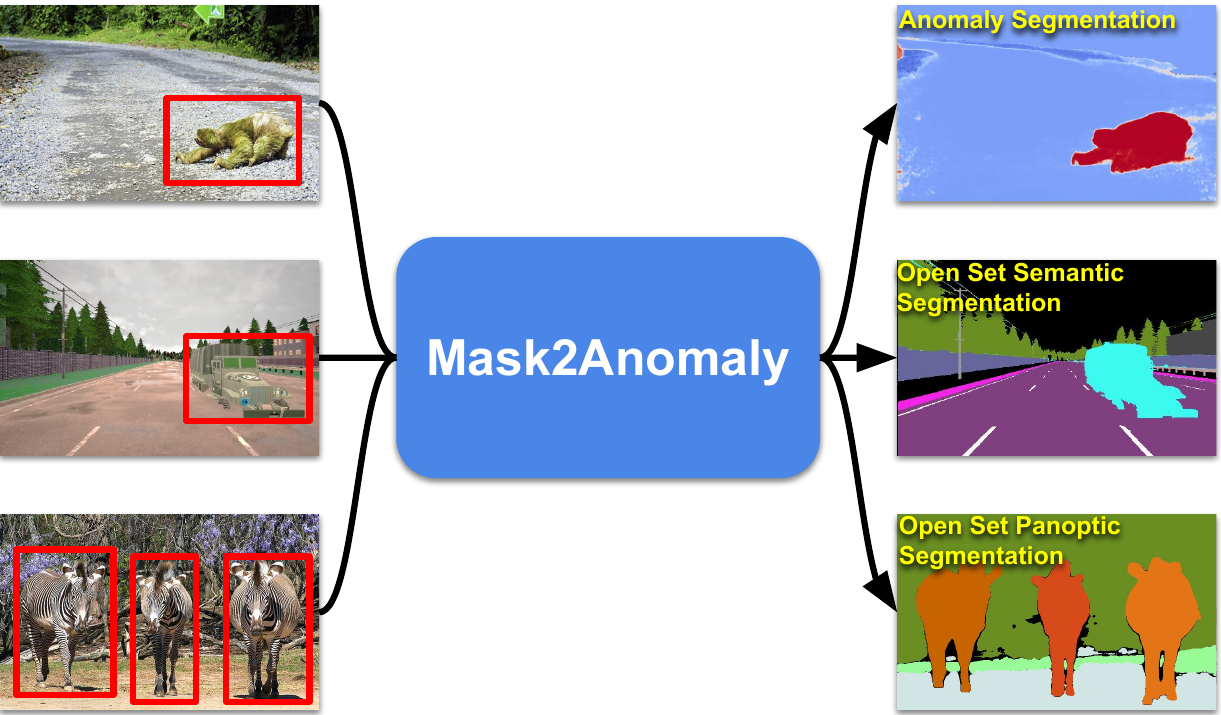}
    \end{center}
    \vspace{-1em}
      \caption{\textbf{\our{}}: We present a mask-based architecture that can jointly perform open-set semantic segmentation, open-set panoptic segmentation, and anomaly segmentation. In the figure, the objects enclosed in red boxes are anomaly/unknown.} 
    \label{fig:teaser} 
\end{figure}

\IEEEPARstart{I}mage segmentation~\cite{Cordts2016Cityscapes,sun2019naae, Zhang2019fsda,yang2020fda,tavera2022pixda} plays a significant role in self-driving cars, being instrumental in achieving a detailed understanding of the vehicle's surroundings. 
Generally, segmentation models are trained to recognize a pre-defined set of semantic classes (e.g., car, pedestrian, road, etc.); however, in real-world applications, they may encounter objects not belonging to such categories (e.g., animals or cargo dropped on the road). Therefore, it is essential for these models to identify objects in a scene that are not present during training i.e., \textit{anomalies}, both to avoid potential dangers and to enable continual learning~\cite{michieli2021continual, cermelli2020modelingthebackground, douillard2020plop, cermelli2022incremental} and open-world solutions~\cite{cen2021deep}.
The segmentation of unseen object categories can be performed at three levels of increasing semantic output information (see \cref{fig:teaser}): 
\begin{itemize}
    \item \emph{Anomaly segmentation} (AS)~\cite{blum2021fishyscapes,xia2020synthesize,fontanel2021detecting,jung2021standardized} focuses on segmenting objects from classes that were absent during training, generating an output map that identifies the anomalous image pixels.
    \item \emph{Open-set semantic segmentation} (OSS)~\cite{grcic2022densehybrid} evaluates a segmentation model's performance on both anomalies and known classes. OSS ensures that when training an anomaly segmentation model, its performance on known classes remains unaffected.
    \item \emph{Open-set panoptic segmentation} (OPS)~\cite{hwang2021exemplar} simultaneously segments distinct instances of unknown objects and performs panoptic segmentation~\cite{kirillov2019panoptic} for the known classes.
\end{itemize}

\begin{figure}[t]
\begin{center}

\includegraphics[width=1\linewidth]{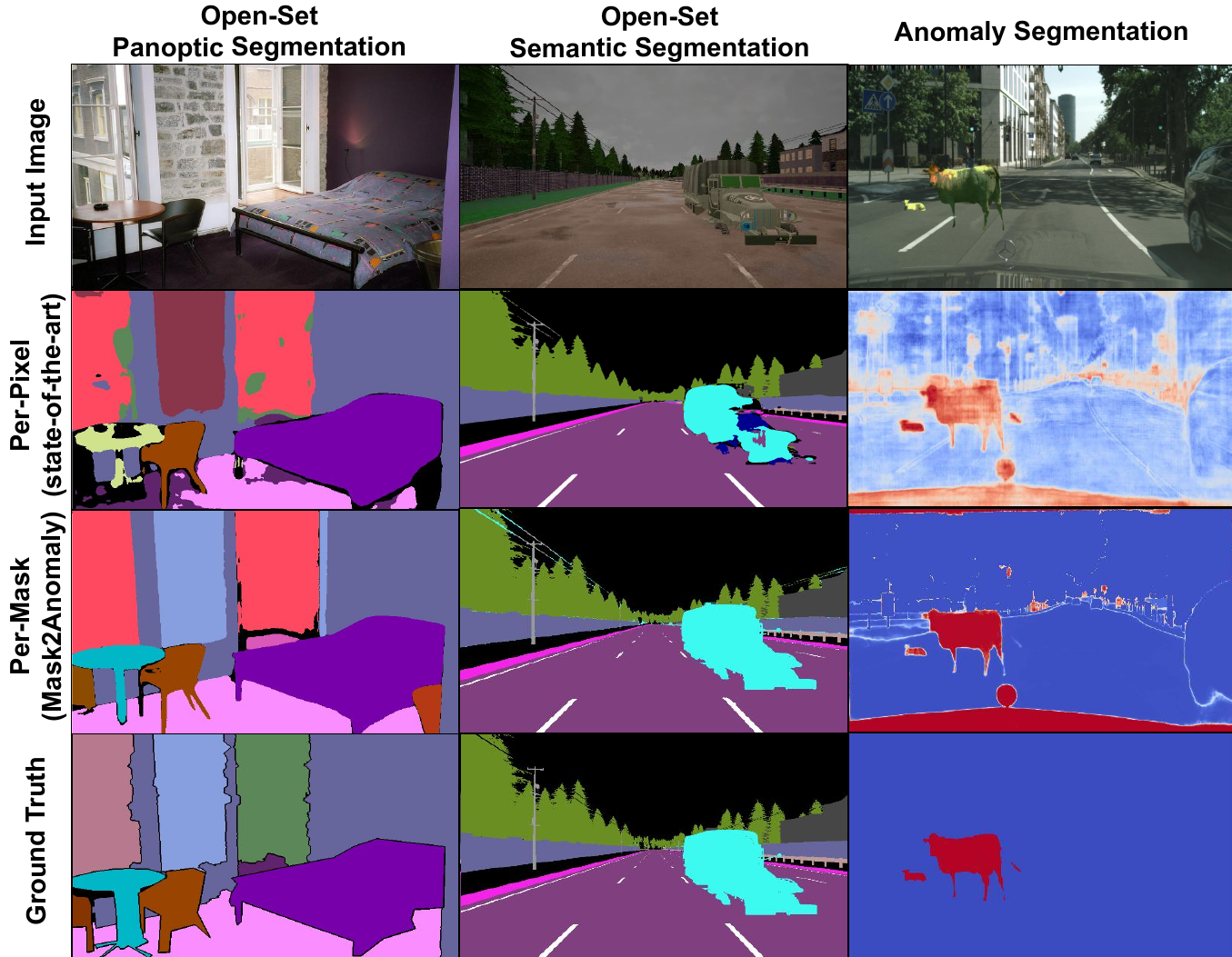}
\end{center}
\vspace{-1em}
  \caption{\textbf{Per-pixel vs per-mask architecture:} We show significant shortcoming in the performance of state-of-the-art methods employing per-pixel architectures for anomaly segmentation or open-set segmentation tasks. These methods prediction have significant false positives and noisy outcomes.~\our{}(ours), an architecture based on mask-transformer properties that effectively addresses both anomaly segmentation and open-set segmentation tasks, leading to a substantial reduction in false positives and enhancing overall prediction quality.} \vspace{-1em}
\label{fig:T2}
\end{figure}

In the literature, AS, OSS and OPS are typically addressed separately using specialized networks for each task.
These networks rely on per-pixel classification architectures that individually classify the pixels and assign to each of them an anomaly score. However, reasoning on the pixels individually without any spatial correlation produces noisy anomaly scores, thus leading to a high number of false positives and poorly localized anomalies or unknown objects (see \cref{fig:T2}).
In this paper, we propose to jointly address AS, OSS, and OPS with a single architecture (with minor changes during inference) by casting them as a mask classification task rather than a pixel classification task (see \cref{fig:teaser}). The idea of employing mask-based architecture stems from the recent advances in mask-transformer architectures~\cite{cheng2022masked, cheng2021per}, which demonstrated that it is possible to achieve remarkable performance across various segmentation tasks by classifying masks rather than pixels. We hypothesize that mask-transformer architectures are better suited to detect anomalies than per-pixel architectures because masks encourage objectness and thus can capture anomalies as whole entities, leading to more congruent anomaly scores and reduced false positives. However, the effectiveness of mask-transformer architectures hinges on the capability to output masks that captures anomalies well. Hence, we propose several technical contributions to improve the capability of mask-transformer architectures to capture anomalies or unknown objects and minimize false positives: 
\begin{itemize}[noitemsep,topsep=0pt]
    \item At the \textbf{architectural} level, we propose a global masked-attention mechanism that allows the model to focus on both the foreground objects and on the background while retaining the efficiency of the original masked-attention~\cite{cheng2022masked}.
    \item At the \textbf{training} level, we have developed a mask contrastive learning framework that utilizes outlier masks from additional out-of-distribution data to maximize the separation between anomalies and known classes.
    \item At the \textbf{inference} level, for anomaly segmentation, we propose a mask-based refinement solution that reduces false positives by filtering masks based on the panoptic segmentation that distinguishes between ``things'' and ``stuff'' and for open-set panoptic segmentation, we developed an approach to mine unknown instances based on mask-architecture properties.
\end{itemize}
We integrate these contributions on top of the mask architecture~\cite{cheng2022masked} and term this solution \textbf{\our}.
To the best of our knowledge, {\our} is the first universal architecture that jointly addresses AS, OSS, and OPS task and segment anomalies or unknown objects at the mask level. We tested {\our} on standard anomaly segmentation benchmarks (Road Anomaly~\cite{lis2019detecting}, Fishyscapes~\cite{blum2021fishyscapes},  Segment Me If You Can~\cite{chan2021segmentmeifyoucan}, Lost\&Found~\cite{pinggera2016lost}), open-set semantic segmentation benchmark (Streethazard~\cite{hendrycks2019scaling}), and  open-set panoptic MS-COCO~\cite{hwang2021exemplar} dataset, achieving the best results among all methods for all task by a significant margin. In particular, {\our} reduces the false positives rate by more than half on average and improves the open-set metric performance by one-third w.r.t the previous state-of-the-art. Code and pre-trained models will be made publicly available upon acceptance.

This work is an extension of our previous paper~\cite{rai2023unmasking} that was accepted to ICCV 2023 (Oral) with the following contributions:
\begin{itemize}[noitemsep,topsep=0pt]
\item We extend~\our{} to open-set segmentation tasks, namely open-set semantic segmentation and open-set panoptic segmentation. 
\item For the open-set panoptic segmentation task, we developed a novel approach to mine unknown instances based on the properties of the mask-architecture and provide related ablation studies to show its efficacy.
\item Extensive qualitative and quantitative experiments demonstrate that~\our{} is an effective approach to address open-set segmentation tasks. Notably,~\our{} gives a significant gain of 30\% on Open-IoU metrics w.r.t best existing method.
\item We extend~\our{} experimentation for the anomaly segmentation task by showing results on the Lost\&Found dataset. Also, we show global mask attention can positively impact semantic segmentation by investigating its generalizability to other datasets. 

\end{itemize}

\section{Related Work}
\label{sec:related}
\myparagraph{Mask-based semantic segmentation.}
Traditionally, semantic segmentation methods \cite{long2015fully, chen2018encoder, zhao2017pyramid, lin2017refinenet, zhang2018exfuse} have adopted fully-convolutional encoder-decoder architectures \cite{long2015fully, badrinarayanan2017segnet} and addressed the task as a dense classification problem. 
However, transformer architectures have recently caused us to question this paradigm due to their outstanding performance in closely related tasks such as object detection~\cite{carion2020end} and instance segmentation~\cite{he2017mask}. In particular, \cite{cheng2021per} proposed a mask-transformer architecture that addresses segmentation as a mask classification problem. It adopts a transformer and a per-pixel decoder on top of the feature extraction. The generated per-pixel and mask embeddings are combined to produce the segmentation output. Building upon~\cite{cheng2021per},~\cite{cheng2022masked} introduced a new transformer decoder adopting a novel masked-attention module and feeding the transformer decoder with one pixel-decoder high-resolution feature at a time. 

So far, all these mask-transformers have been considered exclusively in a closed set setting, i.e, there are no unknown categories at test time. 
To the best of our knowledge, Mask2Anomaly is the first method that performs AS directly with mask-transformers, thus empowering these approaches with the capability to recognize anomalies in real-world settings.

\myparagraph{Anomaly segmentation} methods can be broadly divided into three categories: (a) Discriminative, (b) Generative, and (c) Uncertainty-based methods.~\textit{Discriminative Methods} are based on the classification of the model outputs. Hendrycks and Gimpel~\cite{hendrycks2016baseline} established the initial AS discriminative baseline by applying a threshold over the maximum softmax probability (MSP) that distinguishes between in-distribution and out-of-distribution data. Other approaches use auxiliary datasets to improve performance \cite{liang2017enhancing, jung2021standardized,tian2022pixel} by calibrating the model over-confident outputs.
Alternatively, \cite{lee2018simple} learns a confidence score by using the Mahalanobis distance, and \cite{chan2021entropy} introduces an entropy-based classifier to discover out-of-distribution classes.
Recently, discriminative methods tailored for semantic segmentation \cite{blum2021fishyscapes} directly segment anomalies in embedding space. \textit{Generative Methods } provides an alternative paradigm to segment anomalies based on generative models ~\cite{lis2019detecting, di2021pixel, xia2020synthesize, vojir2021road}. These approaches train generative networks to reconstruct anomaly-free training data and then use the generation discrepancy to detect an anomaly at test time. All the generative-based methods heavily rely on the generation quality and thus experience performance degradation due to image artifacts~\cite{fontanel2021detecting}.
Finally,~\textit{Uncertainty based} methods segment anomalies by leveraging uncertainty estimates via Bayesian neural networks~\cite{mukhoti2018evaluating}. 

\myparagraph{Open-set segmentation} is the task of segmenting both the the anomalies and in-distribution classes for a given image. Anomaly segmentation methods~\cite{hendrycks2018deep,xia2020synthesize} can be adapted to perform open-set semantic segmentation by fusing the in-distribution segmentation results. However, these methods show poor performance in open-set metrics because their in-distribution class segmentation capabilities degrade after training for anomaly segmentation. \cite{bevandic2019simultaneous} formally introduces the problem of open-set semantic segmentation that uses multi-task model segment anomaly and predicts semantic segmentation maps. Later,~\cite{bevandic2021dense} improved the prior method using noisy outlier labels. Recently, \cite{grcic2022densehybrid} proposed a hybrid approach that combines the known class posterior, dataset posterior, and an un-normalized data likelihood to estimate anomalies and in-distribution classes simultaneously. Another challenging problem in the space of open-set segmentation is open-set panoptic segmentation~\cite{hwang2021exemplar}. In open-set panoptic segmentation, the goal is to simultaneously segment  distinct instances of unknown objects and perform panoptic segmentation for in-distribution classes. Hwang~\textit{et.al.}~\cite{hwang2021exemplar} proposed an exemplar-based open-set panoptic segmentation network (EOPSN) that is based on exemplar theory and utilizes Panoptic FPN~\cite{kirillov2019panoptic} which is a per-pixel architecture to perform open-set panoptic segmentation.

All the methods discussed so far for anomaly and open-set segmentation rely on per-pixel classification and evaluate individual pixels without considering local semantics. This approach often leads to noisy anomaly predictions, resulting in significant false positives and reduced in-distribution class segmentation performance. {\our} overcomes this limitation by segmenting anomalies and in-distribution classes as semantically clustered masks, encouraging the objectness of the predictions. To the best of our knowledge, this is the first work to use masks both to segment anomalies and for open-set segmentation.
\section{Preliminaries}
\label{sec:Preliminaries}
\myparagraph{Notations}: Let us denote $\mathcal{X} \subset \mathbb{R}^{3 \times H \times W}$ the space of RGB images, where $H$ and $W$ are the height and width, respectively, and with $\mathcal{Y} \subset \mathbb{N}^{Z \times H \times W}$ the space of semantic labels that associate each pixel in an image to a semantic category from a predefined set $\mathcal{Z}$, with $|\mathcal{Z}|=Z$. At training time we assume to have a dataset $\mathcal{D} = \left\{ (x_i, y_i)\right\}_{i=1}^{D}$, where  $x_i \in \mathcal{X}$ is an image and $y_i\in \mathcal{Y}$ is its ground truth having pixel-wise semantic class labels. Alternatively, $\mathcal{Y}$ can also be described as the semantic partition of the image into $Z$ regions that are represented as a set of binary masks $M^{gt}$, where the ground-truth labels of $x_{i}$ can be represented as $M^{gt} = \{m_i|m_i \in [0,1]^{H\times W}\}_{i=1}^{Z}$.\\

\noindent \textbf{Mask architectures:} The prototypical mask architecture consists of three meta parts: a) a \textit{backbone} that acts as feature extractor, b) a \textit{pixel-decoder} that upsamples the low-resolution features extracted from the backbone to produce high-resolution \textit{per-pixel embeddings}, and c) a \textit{transformer decoder}, made of $L$ transformer layers, that takes the image features to output a fixed number of object queries consisting of \textit{mask embeddings} and their associated \textit{class scores} $C\in \mathbb{R}^{N \times Z}$. The final \textit{class masks} $M\in \mathbb{R}^{N\times (H \times W)}$ are obtained by multiplying the mask embeddings with the per-pixel embeddings obtained from the pixel-decoder. 

During training we use the Hungarian algorithm to  match ground truth masks $M^{gt}$ with the predicted masks $M$. Since the Hungarian algorithm requires one-to-one correspondences and $M\geq M^{gt}$, we pad the ground truth mask $M^{gt}$ with “no object” masks, which we indicate as $\phi$. The  cost function for matching $M$ and $M^{gt}$ is given by
\begin{equation}
    L_{masks} = \lambda_{bce}L_{bce} + \lambda_{dice}L_{dice}
\end{equation}
where $L_{bce}$ and $L_{dice}$ are, respectively, the binary cross entropy loss and the dice loss calculated between the matched masks. The weights $\lambda_{bce}$ and $\lambda_{dice}$ are both set to $5.0$. Additionally, we also train the model on cross-entropy loss $L_{ce}$ to learn the semantic class of each mask that is denoted by $C$. The total training loss is given by:
\begin{equation}
    L = L_{masks} + \lambda_{ce}L_{ce}
\end{equation}
with $\lambda_{ce}$ set to 2.0 for the prediction that is matched with the ground truth and 0.1 for $\phi$, i.e.,  for no object. At inference time, the segmentation output is inferred by marginalization over the softmax of $C$ and sigmoid of $M$ given as:
\begin{equation}
\label{eq:inference}
g(x) = \max^{Z} \left(\text{softmax}(C)^T \cdot \text{sigmoid}(M)\right)
\end{equation}

In the subsequent sections we will address the tasks of anomaly segmentation~(\cref{sec:AS}), open-set semantic segmentation~(\cref{subsec:OSSS}), and open-set panoptic segmentation~(\cref{subsec:OPS}) using our proposed {\our} architecture and delve into its novel elements.
\begin{figure}[t]
\begin{center}
\includegraphics[width=1\linewidth, height=0.6\linewidth]{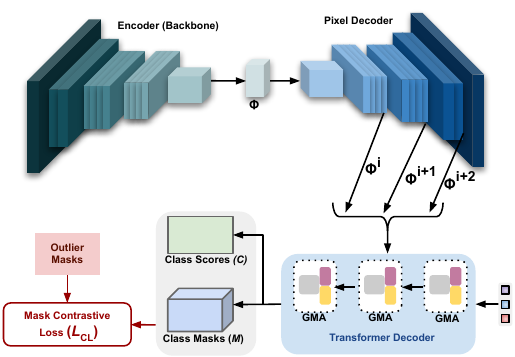}
\end{center}
\vspace{-1em}
  \caption{\textbf{Mask2Anomaly Overview.} Mask2Anomaly meta-architecture consists of an encoder, a pixel decoder, and a transformer decoder. We propose GMA: Global Mask Attention that is discussed in \cref{sec:global_attention} and \cref{fig:gma}. $\phi$ is image features. $\phi^{i}, \phi^{i+1}, \phi^{i+2}$ are upsampled image features at multiple scales. Mask contrastive Loss $L_{CL}$ (\cref{subsec:CL}) utilizes outlier masks to maximize the separation between anomalies and known classes. During anomaly inference, we utilize refinement mask $R_{M}$ (\cref{subsec:RM}) to minimize false positives.} \vspace{-1em} 
\label{fig:overview}
\end{figure}
\section{Anomaly Segmentation}
\label{sec:AS}
\subsection{Problem Setting}
Anomaly segmentation can be achieved in per-pixel semantic segmentation architectures~\cite{chen2018encoder} by applying the \textit{Maximum Softmax Probability} (MSP)~\cite{hendrycks2016baseline} on top of the per-pixel classifier. Formally, given the pixel-wise class scores $S(x) \in [0,1]^{Z\times H \times W}$ obtained by segmenting the image $x$ with a per-pixel architecture, we can compute the anomaly score $f(x)$ as:
\begin{equation}
    \label{eq:perpixel_msp}
    f(x) = 1-\max^{Z}(S(x)).
\end{equation}
In this paper, we propose to adapt this framework based on MSP for mask-transformer segmentation architectures. Given such a mask-transformer architecture, we calculate the 
anomaly scores for an input $x$ as 
\begin{equation}
    \label{eq:mask_msp}
    f(x) = 1 - \max^{Z} \left(\text{softmax}(C)^T \cdot \text{sigmoid}(M)\right).
\end{equation}
Here, $f(x)$ utilizes the same marginalization strategy of
class and mask pairs as~\cite{cheng2021per} to get anomaly scores. 
Without loss of generality, we implement the anomaly scoring (\cref{eq:mask_msp}) on top of the Mask2Former~\cite{cheng2022masked} architecture. However, this strategy hinges on the fact that the masks predicted by the segmentation architecture can capture anomalies well. We found that simply applying the MSP on top of Mask2Former as in \cref{eq:mask_msp} does not yield good results (see \cref{fig:teaser} and the results in \cref{sec:ablation}). To overcome this problem, we introduce improvements in the architecture, training procedure, and anomaly inference mechanism. We name our method as {\our},  and its overview is shown in \cref{fig:overview} (left). Now, we will discuss the proposed novel components in~\our{}.
\begin{figure}[t]
\begin{center}
\includegraphics[width=1\linewidth]{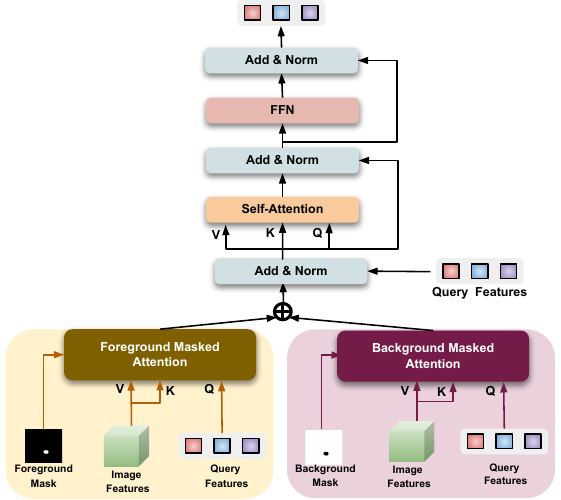}
\end{center}
\vspace{-1em}
  \caption{\textbf{Global Mask Attention:} independently distributes the attention between foreground and background. V, K, and Q are Value, Key, and Query.  } \vspace{-1em}
\label{fig:gma}
\end{figure}
\subsection{Global Masked Attention}
\label{sec:global_attention}
One of the key ingredients to Mask2Former~\cite{cheng2022masked} state-of-the-art segmentation results is the replacement of the \textit{cross-attention} (CA) layer in the transformer decoder with a \textit{masked-attention} (MA). The masked-attention attends only to pixels within the foreground region of the predicted mask for each query, under the hypothesis that local features are enough to update the query object features. 
The output of the $l$-th masked-attention layer can be formulated as
\begin{equation}
    \text{softmax}(\mathcal{M}^F_l + QK^{T})V + X_{in} 
\end{equation}
where $X_{in}\in$ $\mathbb{R}^{N \times C}$ are the $N$ $C$-dimensional query features from the previous decoder layer. The queries $Q \in \mathbb{R}^{N \times C}$ are obtained by linearly transforming the query features with a learnable transformation whereas the keys and values $K, V$ are the image features under learnable linear transformations $f_k(.)$ and $f_v()$. 
Finally, $\mathcal{M}_l^F$ is the predicted foreground attention mask that at each pixel location $(i, j)$ is defined as
\begin{equation}
    \mathcal{M}_l^F(i, j) = \begin{cases}
        0 & \text{if } M_{l-1}(i, j)  \geq 0.5 \\
        -\infty & \text{otherwise},
    \end{cases}
\end{equation}
where $M_{l-1}$ is the output mask of the previous layer. 

\begin{figure}[t]
    \begin{center}
        \includegraphics[width=1\linewidth]
        {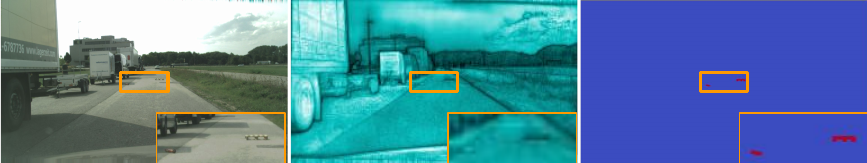}
    \end{center}
    \vspace{-0.75em}
    {\small{\hspace{2.25em}Input Image \hspace{2.75em}Attention Map\hspace{3.25em}Ground Truth}}
\caption{\textbf{Limitation of Mask-Attention:} Masked-attention~\cite{cheng2022masked}
selectively attends to foreground regions resulting in low attention scores (dark regions) for anomalies. Anomalies are in~{red}. Best viewed with Zoom.}
    \label{fig:mask2former_attention}
    \vspace{-1em}
\end{figure}

By focusing only on the foreground objects, masked attention grants faster convergence and better semantic segmentation performance than cross-attention. However, focusing only on the foreground region constitutes a problem for anomaly segmentation because anomalies may also appear in the background regions. Removing background information leads to failure cases in which the anomalies in the background are entirely missed, as shown in the example in \cref{fig:mask2former_attention}. To ameliorate the detection of anomalies in these corner cases, we extend the masked attention with an additional term focusing on the background region (see \cref{fig:gma}, right). We call this a \textit{global masked-attention} (GMA) formally expressed as 
\begin{equation}
    \label{eq:gma}
    \begin{aligned}
        X_{out} = &\text{softmax}(\mathcal{M}_l^{F} + QK^{T})V \\
        +& \text{softmax}(\mathcal{M}_l^{B} + QK^{T})V + X_{in}
    \end{aligned}
\end{equation}
where $ \mathcal{M}_l^B$ is the additional background attention mask that complements the foreground mask $\mathcal{M}_l^F$, and it is defined at the pixel coordinates $(i, j)$ as
\begin{equation}
    \mathcal{M}_l^B(i, j) = \begin{cases}
        0 & \text{if } M_{l-1}(i, j)  < 0.5 \\
        -\infty & \text{otherwise}.
    \end{cases}
\end{equation}

The global masked-attention in \cref{eq:gma} differs from the masked-attention by additionally attending to the background mask region, yet it retains the benefits of faster convergence w.r.t. the cross-attention. 
\subsection{Mask Contrastive Learning}
\label{subsec:CL}
The ideal characteristic of an anomaly segmentation model is to predict high anomaly scores for out-of-distribution (OOD) objects and low anomaly scores for in-distribution (ID) regions. Namely, we would like to have a significant margin between the likelihood of known classes being predicted at anomalous regions and vice-versa.
A common strategy used to improve this separation is to fine-tune the model with auxiliary out-of-distribution (anomalous) data as supervision~\cite{grcic2020dense,grcic2022densehybrid, blum2021fishyscapes}.  

Here we propose a contrastive learning approach to encourage the model to have a significant margin between the anomaly scores for in-distribution and out-of-distribution classes. 
Our mask-based framework allows us to straightforwardly implement this contrastive strategy by using as supervision 
outlier images generated by cutting anomalous objects from the auxiliary OOD data and pasting it on top of the training data. For each outlier image, we can then generate a binary outlier mask $M_{OOD}$ that is $1$ for out-of-distribution pixels and $0$ for in-distribution class pixels.
With this setting, we first calculate the negative likelihood of in-distribution classes using the class scores $C$ and class masks $M$ as:
\begin{equation}
    \begin{aligned}
         l_{N} = - \max^{Z} \left(\text{softmax}(C)^T \cdot \text{sigmoid}(M)\right)
    \end{aligned}
\end{equation}
Ideally,  for pixels corresponding to in-distribution classes $l_{N}$ should be $-1$ since the value of $\text{softmax}(C)^T$ and $\text{sigmoid}(M)$ would be close to $1$. On the other hand, for the anomalous pixels, $l_{N}$ should be $0$ as the likelihood of these pixels belonging to any in-distribution classes is $0$ resulting $\text{softmax}(C)^T$ to be $0$. Using $l_{N}$, we define our contrastive loss as:
\begin{equation}
    \begin{aligned}
         L_{CL} &= \frac{1}{2}(l_{CL}^{2}),\\
          l_{CL} &= 
    \begin{cases}
        l_{N} & \text{if} M_{OOD} = 0 \\
        max(0, m - l_{N}) & \text{otherwise,}
    \end{cases}
    \end{aligned}
\end{equation}
where the margin $m$ is a hyperparameter that decides the minimum distance between the out-of-distribution and in-distribution classes. During mask contrastive training, we also preserve the in-distribution accuracy by training on $L_{masks}$ and $L_{ce}$ which formulates our total training loss as:
\begin{equation}
    L_{ood} = L_{CL} + L_{masks} + \lambda_{ce}L_{ce}
\end{equation}
\begin{figure}[t]
    \begin{center}
        \includegraphics[width=1\linewidth]{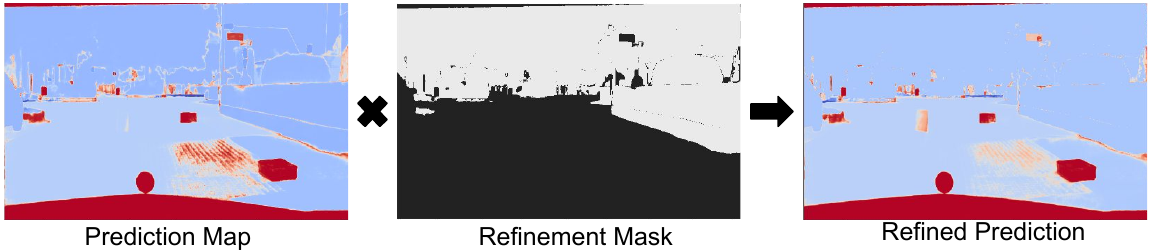}
    \end{center}
    \vspace{-1em}
    \caption{\textbf{Mask Refinement Illustration:} To obtain the refined prediction, we multiply the prediction map with a refinement mask that is built by assigning zero anomaly scores for pixels that are categorized as ``stuff'', except for the ``road''. The refinement eliminates many false positives at the boundary of objects and in the background. The region to be masked is white in the refinement mask.}
    \vspace{-1em}
    \label{fig:refinements}
\end{figure}
\subsection{Refinement Mask}
\label{subsec:RM}
False positives are one of the main problems in anomaly segmentation, particularly around object boundaries. Handcrafted methods such as iterative boundary suppression ~\cite{jung2021standardized} or dilated smoothing have been proposed to minimize the false positives at boundaries or globally, however, they require tuning for each specific dataset. 
Instead, we propose a general 
refinement technique that leverages the capability of mask transformers~\cite{cheng2022masked} to perform all segmentation tasks.
Our method stems from the panoptic perspective~\cite{kirillov2019panoptic} that the elements in the scene can be categorized as \textit{things}, i.e. countable objects, and \textit{stuff}, i.e. amorphous regions. 
With this distinction in mind, we observe that in driving scenes, i) unknown objects are classified as things, and ii) they are often present on the road. Thus, we can proceed to remove most false positives by filtering out all the masks corresponding to “stuff", except the ``road" category. We implement this removal mechanism in the form of a binary refinement mask $R_{M} \in [0,1]^{H \times W}$, which contains zeros in the segments corresponding to the unwanted ``stuff" masks and one otherwise. Thus, by multiplying $R_{M}$ with the predicted anomaly scores $f$ we filter out all the unwanted ``stuff" masks and eliminate a large portion of the false positives (see \cref{fig:refinements}). Formally, for an image $x$ the refined anomaly scores $f^r$ is computed as:
\begin{equation}
\label{eq:refinemnt_mask}
    f^r(x) = R_{M} \odot f(x),
\end{equation}
where $\odot$ is the Hadamard product. 

$R_M$ is the dot product between the binarized output mask $\bar{M} \in \{0,1\}^{N \times (H \times W)}$ and the class filter $\bar{C} \in \{0,1\}^{1 \times N}$, i.e. $R_M = \bar{C} \cdot \bar{M}$.
We define $\bar{M} = \text{sigmoid}(M) > 0.5$ and the class filter $\bar{C}$ is equal to $1$ only where the highest class score of $\text{softmax}(C)$ belongs to ``things" or ``road" classes and is greater than $0.95$. 

\myparagraph{Inference:} During inference, we pass the input image through~\our{} to get anomaly scores \cref{eq:mask_msp}. Then, we refine the anomaly scores via refinement mask \cref{eq:refinemnt_mask}.

\section{Open-Set Semantic Segmentation}
\label{subsec:OSSS}
\subsection{Problem Setting} Anomaly segmentation methods solely focus on segmenting road scene anomalies. However, a strong performance for in-distribution classes is equally important. For instance, an anomaly segmentation model deployed in an autonomous vehicle that fails to identify a person crossing the road can result in a fatal accident. Hence, it is crucial that while recognizing anomalies, the  performance of the model on in-distribution classes remains preserved. 
Open-set semantic segmentation addresses this problem by jointly accessing the model's performance on in-distribution and out-of-distribution classes. We utilize the mask properties of~\our{} to perform open-set semantic segmentation by only modifying its inference process with respect to the Anomaly Segmentation task.

\myparagraph{Inference:} Our open-set semantic segmentation network has an identical mask architecture as anomaly segmentation that contains global mask attention. During the inference, we first threshold the anomaly scores obtained from~\cref{eq:mask_msp} at a true positive rate of 95\%, similar to ~\cite{grcic2022densehybrid}. We denote the thresholded anomaly scores by $\hat{f(x)}$. Next, we calculate the in-distribution class performance $g(x)$ by~\cref{eq:inference}. Finally, we formulate the open-set semantic segmentation $f_{oss}$ prediction of an image $x$ as: 
\begin{equation}
    f_{oss}(x) = \arg \max_{} (\text{concat}(g(x), \hat{f(x)}))
\end{equation}
\section{Open Set Panoptic Segmentation}
\label{subsec:OPS}
\subsection{Problem Overview} Panoptic segmentation~\cite{kirillov2019panoptic} jointly addresses the dense prediction task of semantic segmentation and instance segmentation. In this task, we divide an image into two broad categories: i) \textit{stuff}, i.e., amorphous areas of an image that have homogeneous texture, such as grass and sky, and ii) \textit{things}, i.e., countable objects such as pedestrians. Every pixel belonging to a \textit{things} category is assigned a semantic label and a unique instance id, whereas, for \textit{stuff} regions, only semantic labels are given, and the instance id is ignored. However, constructing and annotating large-scale panoptic segmentation datasets is expensive and requires significant human effort. Hwang~\cite{hwang2021exemplar} addresses this problem by formulating it as open-set panoptic segmentation (OPS) problem where a model can perform panoptic segmentation on a pre-defined set of classes and identify unknown objects. This ability of the OPS model could accelerate the process of constructing large-scale panoptic segmentation datasets from existing ones.
\subsection{Problem Setting} The key difference between panoptic and open-set panoptic segmentation is the presence of unknown objects while testing. However, handling the classification of unknown object is OPS is quite challenging. Firstly, in comparison to open-set image classification, OPS requires the classification of unknown objects at the pixel level. Secondly, the absence of semantic information about unknown objects means that they are generically labeled as background during training. 
In order to make the problem tractable, we follow~\cite{hwang2021exemplar} and make three assumptions: 
\begin{enumerate}
    \item we categorize all the unknowns into things categories (i.e., the unknowns are countable objects);
    \item elements of known categories cannot be classified as unknown classes;
    \item the unknown objects are always found in the background/void regions. This avoids confusion between known and unknown class regions.
\end{enumerate}

We address open-set panoptic segmentation by utilizing the mask properties of~\our{} and leveraging its global mask attention. We first mask out the known \textit{stuff} and \textit{things} regions of an image, and then within the remaining background area, we mine the instances of the unknown objects. We will now formally discuss the method in more detail. For an input image $x$, \our{} outputs a set of masks $M$ and its corresponding class scores $C$. Among these, we denote the joint set of known~\textit{stuff} and~\textit{things} class masks as $M_{k} \in [0,1]^{N_{k} \times H\times W} $ and its corresponding class scores as $C_{k}\in \mathbb{R}^{N_{k} \times Z}$. Finally, we denote the number of known class masks as $N_{k}$. 
We obtain the background region $\mathcal{B}$ of $x$ by using the weighted combination of $M_{k}$ and $C_{k}$ given by:
\begin{equation}
\label{equation_bkg}
   \mathcal{B}  = 1 - \max^{N_{k}}( \max^{Z}(\text{softmax}(C_{k})) \cdot \text{sigmoid}(M_{k}) ) .
\end{equation}
In light of our assumptions, $\mathcal{B}$ consists of background~\textit{stuff} classes and unknown~\textit{things} classes.

\subsection{Mining Unknown Instances} Generally in panoptic segmentation datasets such as MS-COCO~\cite{lin2014microsoft} the background class consists of only background~\textit{stuff} classes. However, in open-set panoptic segmentation, the background class consists of background~\textit{stuff} classes and unknown~\textit{things} classes. So, we mine the unknown instances from background $\mathcal{B}$ obtained from~\cref{equation_bkg} using the following steps: 

\begin{enumerate}[noitemsep,topsep=0pt]
\item In the first step, we employ the connected component algorithm \cite{bieniek1998connected} to cluster and identify unique segments in $\mathcal{B}$.
\item Next, we calculate each connected component's overlap with the individual masks of $M$. Intersection over union is used for calculating the overlap. 
\item If there is a significant overlap between a connected component and a mask $M^{i} \in M$, we calculate the average stuff class entropy $\mathcal{E}_{S}$ and average things class entropy $\mathcal{E}_{T}$ using the corresponding class scores $C^{i} \in C$.
\item Finally, if $\mathcal{E}_{S} > \mathcal{E}_{T}$ we can conclude that the connected component is more likely to belong to the~\textit{things} class. Hence, we classify the connected component to be an unknown instance. 
\end{enumerate}
\myparagraph{Inference:} During the inference, we first calculate $\mathcal{B}$ from~\cref{equation_bkg}. Then, we identify the unknown instances in $\mathcal{B}$ by following the above described steps of mining unknown instances.

\section{Experimentation}
\subsection{Datasets} 
\myparagraph{Anomaly Segmentation}: We train Mask2Anomaly on the Cityscapes~\cite{Cordts2016Cityscapes} dataset, which consists of 2975 training and 500 validation images. To evaluate anomaly segmentation, we use Road Anomaly~\cite{lis2019detecting}, Lost \& Found~\cite{pinggera2016lost}, Fishyscapes~\cite{blum2021fishyscapes}, and Segment Me If You Can (SMIYC) benchmarks~\cite{chan2021segmentmeifyoucan}. \noindent\textit{Road Anomaly:} is a collection of 60 web images with anomalous objects on or near the road. \noindent\textit{Lost \& Found:} has 1068 test images with small obstacles for road scenes. \noindent\textit{Fishyscapes (FS):} consists of two datasets, Fishyscape static (FS static) and Fishyscapes lost \&  found (FS lost \&  found). Fishyscapes static is built by blending Pascal VOC~\cite{everingham2010pascal} objects on Cityscapes images containing 30 validation and 1000 test images. Fishyscapes lost \& found is based on a subset of the Lost and Found dataset~\cite{pinggera2016lost}, with 100 validation and 275 test images. \noindent\textit{SMIYC:} consists of two datasets, RoadAnomaly21 (SMIYC-RA21) and RoadObstacle21 (SMIYC-RO21). The SMIYC-RA21  contains 10 validation and 100 test images with diverse anomalies. The SMIYC-RO21 is collected to segment road anomalies and has 30 validation and 327 test images. 

\myparagraph{Open-set panoptic segmentation:} We perform all the open-set panoptic segmentation experiments on the panoptic segmentation dataset of MS-COCO~\cite{lin2014microsoft}. The dataset consists of 118 thousand training images and 5 thousand validation images having 80 \textit{thing} classes and 53 \textit{stuff} classes. We construct open-set panoptic segmentation dataset by removing the labels of a small set of known \textit{things} classes from the train set of panoptic segmentation dataset. The removed set of \textit{things} classes are treated as unknown classes. We construct three different training dataset split with increasing order of difficulty with (5\%, 10\%, 20\%) of unknown classes. The removed classes in each split that are removed cumulatively is given as: 5\%: \{car, cow, pizza, toilet\}, 10\%: \{boat, tie, zebra, stop sign \}, 20\%: \{dining table, banana, bicycle, cake, sink, cat, keyboard, bear\}.

\myparagraph{Open-set semantic segmentation:} We use StreetHazards~\cite{hendrycks2019scaling}, a synthetic dataset for open-set semantic segmentation. StreetHazards dataset is created with the CARLA simulator~\cite{dosovitskiy2017carla}, leveraging the Unreal Engine to render realistic road scene images in which diverse anomalous objects are inserted. The dataset consists of 5125 training images and 1031 validation images having 12 classes. The test set has 1500 images along with an additional anomaly class.
\subsection{Evaluation Metrics}
\myparagraph{Anomaly Segmentation:} We evaluate all the anomaly segmentation methods at pixel and component levels that are described next.

\noindent\underline{\emph{Pixel-Level}:}  For pixel-wise evaluation, $Y \in \{ Y_{a}, Y_{na}\}$ is the pixel level annotated ground truth labels for an image $\chi$ containing anomalies. $Y_{a}$ and $Y_{na}$ represents the anomalous and non-anomalous labels in the ground-truth, respectively. Assume that $\hat{Y}(\gamma)$ is the model prediction obtained by thresholding at $\gamma$. Then, we can write the precision and recall equations as
\begin{equation}
\text{precision}(\gamma) = \frac{|Y_{a} \cap \hat{Y}_{a}(\gamma)|} {|\hat{Y}_{a}(\gamma)|}
\end{equation}

\begin{equation}
\text{recall}(\gamma) = \frac{|Y_{a} \cap \hat{Y}_{a}(\gamma)|}{|Y_{a}|}
\end{equation}

\noindent and the AuPRC can be approximated as

\begin{equation}
\text{AuPRC} = \int_{\gamma}^{}\text{precision}(\gamma)\text{recall}(\gamma)
\end{equation}
The AuPRC works well for unbalanced datasets making it particularly suitable for anomaly segmentation since all the datasets are significantly skewed. Next, we consider the False Positive Rate at a true positive rate of 95\% (FPR$_{95}$), an important criterion for safety-critical applications that is calculated as:
\begin{equation}
\text{FPR}_{95} = \frac{|\hat{Y}_{a}(\gamma^{*}) \cap Y_{na}|}{|Y_{na}|}
\end{equation}
where $\gamma^{*}$ is a threshold when the true positive rate is 95\%.

\noindent\underline{\emph{Component-Level}:} SMIYC~\cite{chan2021segmentmeifyoucan} introduced component-level evaluation metrics that solely focus on detecting anomalous objects regardless of their size. These metrics are important to be considered because pixel-level metrics may not penalize a model for missing a small anomaly, even though such a small anomaly may be important to be detected. In order to have a component-level assessment of the detected anomalies, the quantities to be considered are the component-wise true-positives ($TP$), false-negatives ($FN$), and false-positives ($FP$). These component-wise quantities can be measured by considering the anomalies as the positive class. From these quantities, we can use three metrics to evaluate the component-wise segmentation of anomalies: sIoU, PPV, and F1$^{*}$. Here we provide the details of how these metrics are computed, using the notation $\mathcal{K}$ to denote the set of ground truth components, and $\hat{\mathcal{K}}$ to denote the set of predicted components.

The \textit{sIoU} metric used in SMIYC~\cite{chan2021segmentmeifyoucan} is a modified version of the component-wise intersection over union proposed in~\cite{rottmann2020prediction}, which considers the ground-truth components in the computation of the $TP$ and $FN$. Namely, it is computed as 
\begin{equation}
    \text{sIoU}(k) = \frac{|k\cap \hat{K}(k)|}{|k\cap \hat{K}(k) \backslash \mathcal{A}(k)|},
    \qquad \hat{K}(k) = \bigcup_{\hat{k} \in \hat{\mathcal{K}}, \, \hat{k} \cap k \neq \emptyset} \hat{k}
\end{equation}
where $\mathcal{A}(k)$ is an adjustment term that excludes from the union those pixels that correctly intersect with another ground-truth component different from $k$.
Given a threshold $\tau \in [0, 1]$, a target $k \in \mathcal{K}$ is considered a $TP$ if $sIoU(k) > \tau$, and a $FN$ otherwise.

The positive predictive value (\textit{PPV}) is a metric that measures the $FP$ for a predicted component $\hat{k} \in \hat{\mathcal{K}}$, and it is computed as 
\begin{equation}
    \text{PPV}(\hat{k}) = \frac{|\hat{k}\cap \hat{K}(k)|}{|\hat{k}|}
\end{equation}
A predicted component $\hat{k} \in \hat{\mathcal{K}}$ is considered a $FP$ if $PPV(\hat{k}) \leq \tau$.
Finally, the \textit{$F1^{*}$} summarizes all the component-wise $TP$, $FN$, and $FP$ quantities by the following formula:
\begin{equation}
    \label{f1*}
    F1^{*}(\tau) = \frac{2TP(\tau)}{2TP(\tau) + FN(\tau) + FP(\tau)}
\end{equation}
\myparagraph{Open-set semantic segmentation:} We use open-IoU~\cite{grcic2022densehybrid} to evaluate open-set semantic segmentation. Unlike, IoU, open-IoU takes into account the false positives ($FP^{OOD}$) and false negatives ($FN^{OOD}$) of an anomaly segmentation model. To measure open-IoU, we first threshold the output of the anomaly segmentation model at a true positive rate of 95\% and then re-calculate the classification scores of in-distribution classes according to the anomaly threshold. Now, $FP^{OOD}$ and $FN^{OOD}$ for a class $\alpha$ can be calculated as:
\begin{equation}
FP_{\alpha}^{OOD} = \sum_{i=1,i\neq \alpha}^{Z + 1} FP_{\alpha}^{i},  FN_{\alpha}^{OOD} = \sum_{i=1,i\neq \alpha}^{Z + 1} FN_{\alpha}^{i}
\end{equation}
Using $FP_{\alpha}^{OOD}$ and $FN_{\alpha}^{OOD}$, we can calculate the open-IoU for class $\alpha$ as: 
\begin{equation}
\text{open-IoU}_{\alpha} = \frac{TP_{\alpha}}{TP_{\alpha} + FP_{\alpha}^{OOD} + FN_{\alpha}^{OOD}}
\end{equation}
$TP_{\alpha}$ denotes the true-positive of class $\alpha$. An ideal open-set model will have open-IoU to be equal to IoU.

\myparagraph{Open-set panoptic segmentation:} We measure the panoptic segmentation quality of known and unknown classes by using the panoptic quality ($PQ$) metric~\cite{kirillov2019panoptic}. For each class, $PQ$ is calculated individually and averaged over all the classes making $PQ$ independent of class imbalance. Every class has predicted segments $p$ and its corresponding ground truths $g$ that is divided into three parts: true positives ($TP$): matched pair of segments, false positives ($FP$): unmatched predicted segments, and false negatives ($FN$): unmatched ground truth segments. Given the three sets, $PQ$ can be formulated as:
\begin{equation}
    PQ = \underbrace{\frac{ \sum_{(p,g) \in TP } IoU(p,g)}{|TP|}}_\text{segmentation quality (SQ)} \times \underbrace{\frac{|TP|}{|TP| + \frac{1}{2}|FP| + \frac{1}{2}|FN|}}_\text{recognition quality (RQ)}
\end{equation}
From the above equation, we can see $PQ$ as the product of a segmentation quality ($SQ$) and a recognition quality ($RQ$). $RQ$ can be inferred as an F1 score that gives the estimation of segmentation quality. $SQ$ is the average IoU of matched segments.
\subsection{Implementation Details}  
\myparagraph{Anomaly Segmentation:} Our implementation is derived from~\cite{cheng2021per,cheng2022masked}. We use a ResNet-50~\cite{he2016deep} encoder, and its weights are initialized from a model that is pre-trained with barlow-twins~\cite{zbontar2021barlow} self-supervision on ImageNet~\cite{deng2009imagenet}. We freeze the encoder weights during training, saving memory and training time. We use a multi-scale deformable attention Transformer (MSDeformAttn)~\cite{zhu2020deformable} as the pixel decoder. The MSDeformAttn gives features maps at $1/8, 1/16,$ and $1/32$ resolution, providing image features to the transformer decoder layers. Our transformer decoder is adopted from~\cite{cheng2022masked} and consists of 9 layers with 100 queries. We train~\our{} using a combination of binary cross-entropy loss and the dice loss~\cite{milletari2016v} for class masks and cross-entropy loss for class scores.  The network is trained with an initial learning rate of 1e-4 and batch size of 16 for 90 thousand iterations on AdamW~\cite{loshchilov2017decoupled} with a weight decay of 0.05. We use an image crop of $380\times760$ with large-scale jittering~\cite{du2021simple} along with a random scale ranging from 0.1 to 2.0. Next, we train the~\our{} in a contrastive setting. We generate the outlier image using AnomalyMix~\cite{tian2022pixel} where we cut an object from MS-COCO~\cite{lin2014microsoft} dataset image and paste them on the Cityscapes image. The corresponding binary mask for an outlier image is created by assigning $1$ to the MS-COCO image area and $0$ to the Cityscapes image area. We randomly sample 300 images from the MS-COCO dataset during training to generate outliers. We train the network for 4000 iterations with $m$ as 0.75, a learning rate of 1e-5, and batch size 8, keeping all the other hyper-parameters the same as above. The probability of choosing an outlier in a training batch is kept at 0.2. 
\begin{table*}[t]
\centering
\renewcommand{\arraystretch}{1.15}
\resizebox{1\linewidth}{!}{\begin{tabular}{c|cc|cc|cc|cc|cc|cc|cc}
\multicolumn{1}{c|}{}&\multicolumn{2}{c|}{SMIYC RA-21}&\multicolumn{2}{c|}{SMIYC RO-21}&\multicolumn{2}{c|}{FS L\&F}&\multicolumn{2}{c|}{FS Static}&\multicolumn{2}{c|}{Road Anomaly}& \multicolumn{2}{c|}{Lost \& Found}& \multicolumn{2}{c}{Average}\\
\cline{2-3}\cline{4-5}\cline{6-7}\cline{8-9}\cline{10-11}\cline{12-13}\cline{14-15}
\multicolumn{1}{c|}{Methods}&\multicolumn{1}{c|}{AuPRC $\uparrow$}&\multicolumn{1}{c|}{FPR$_{95}$ $\downarrow$}&\multicolumn{1}{c|}{AuPRC $\uparrow$} &\multicolumn{1}{c|}{FPR$_{95}$ $\downarrow$}&\multicolumn{1}{c|}{AuPRC $\uparrow$} &\multicolumn{1}{c|}{FPR$_{95}$ $\downarrow$}&\multicolumn{1}{c|}{AuPRC $\uparrow$} &\multicolumn{1}{c|}{FPR$_{95}$ $\downarrow$}&\multicolumn{1}{c|}{AuPRC $\uparrow$} &\multicolumn{1}{c|}{FPR$_{95}$ $\downarrow$}&\multicolumn{1}{c|}{AuPRC $\uparrow$} &\multicolumn{1}{c}{FPR$_{95}$$\downarrow$}&\multicolumn{1}{c|}{AuPRC $\uparrow$} &\multicolumn{1}{c}{FPR$_{95}$$\downarrow$}\\
\Xhline{3\arrayrulewidth}
Max Softmax~\cite{hendrycks2016baseline}(ICLR'17) &27.97	&72.02	&15.72	&16.6	&1.77	&44.85	&12.88	&39.83	&15.72	&71.38	&30.14	&33.20 &17.36 &46.31\\
Entropy~\cite{hendrycks2016baseline}(ICLR'17) &-	&-	&-	&-	&2.93	&44.83	&15.4	&39.75	&16.97	&71.1	&- &- &11.76 &51.89\\
Mahalanobis~\cite{lee2018simple}(NeurIPS'18) &20.04	&86.99	&20.9	&13.08	&-	&-	&-	&-	&14.37	&81.09 &54.97 &12.89 &27.57 &48.51\\

Image Resynthesis~\cite{lis2019detecting}(ICCV'19) &52.28	&25.93	&37.71	&4.7&	5.7	&48.05	&29.6	&27.13	&-	&- &57.08 &8.82 &36.47 &22.92 \\
Learning Embedding~\cite{blum2021fishyscapes}(IJCV'21) &37.52	&70.76	&0.82	&46.38	&4.65	&24.36	&57.16	&13.39	&-	&- &61.70 &10.36 &32.37 &33.05\\

Void Classifier~\cite{blum2021fishyscapes}(IJCV'21)&36.61	&63.49	&10.44	&41.54	&10.29	&22.11	&4.5	&19.4	&-	&-	&4.81	&47.02 &13.33 &38.71\\

JSRNet~\cite{vojir2021road}(ICCV'21)&33.64	&43.85&	28.09	&28.86	&-	&-	&-	&-	&\textbf{94.4}	&\textbf{9.2}	&74.17	&6.59 &57.57 &22.12\\
SML~\cite{jung2021standardized}(ICCV'21) &46.8	&39.5	&3.4	&36.8	&31.67	&21.9	&52.05	&20.5	&17.52	&70.7 &- &-&30.28 &37.88\\
SynBoost~\cite{di2021pixel}(CVPR'21) &56.44	&61.86	&71.34	&3.15	&43.22	&15.79	&72.59	&18.75	&38.21	&64.75 &81.71 &\underline{4.64} &60.58 &28.15 \\
Maximized Entropy~\cite{chan2021entropy}(ICCV'21) &\underline{85.47}	&15.00	&85.07	&0.75	&29.96	&35.14	&86.55	&8.55	&48.85	&31.77	&77.90	&9.70 &\underline{68.96} &16.81 \\
Dense Hybrid~\cite{grcic2022densehybrid}(ECCV'22)&77.96&	\textbf{9.81}	&\underline{87.08}	&\underline{0.24}	&\textbf{47.06}	&\textbf{3.97}&	80.23	&5.95	&31.39	&63.97 & 78.67 &\textbf{2.12} &67.06 &\underline{14.34}\\
PEBEL~\cite{tian2022pixel}(ECCV'22)&49.14	&40.82	&4.98	&12.68	&44.17	&7.58	&\underline{92.38}	&\underline{1.73}	&45.10	&44.58	&57.89	&4.73 &48.94 &18.68\\
\hline
\textbf{Mask2Anomaly (Ours)}& \textbf{88.70}	&\underline{14.60}	&\textbf{93.30}	&\textbf{0.20}	&\underline{46.04}	&\underline{4.36}	&\textbf{95.20}	&\textbf{0.82}	&\underline{79.70}	&\underline{13.45}	&\textbf{86.59}	&5.75 &\textbf{81.59} &\textbf{6.53}\\
\end{tabular}}
\caption{\textbf{Pixel level evaluation:} On average, Mask2Anomaly shows significant improvement  among the compared methods. Higher values for AuPRC are better, whereas for FPR$_{95}$ lower values are better. The best and second best results are \textbf{bold} and \underline{underlined}, respectively. `-' indicates the unavailability of benchmark results. } \vspace{-1em}
\label{tab:main} 
\end{table*}

\begin{figure}[t]
    \begin{center}
        \includegraphics[width=1\linewidth, height=0.75\linewidth]{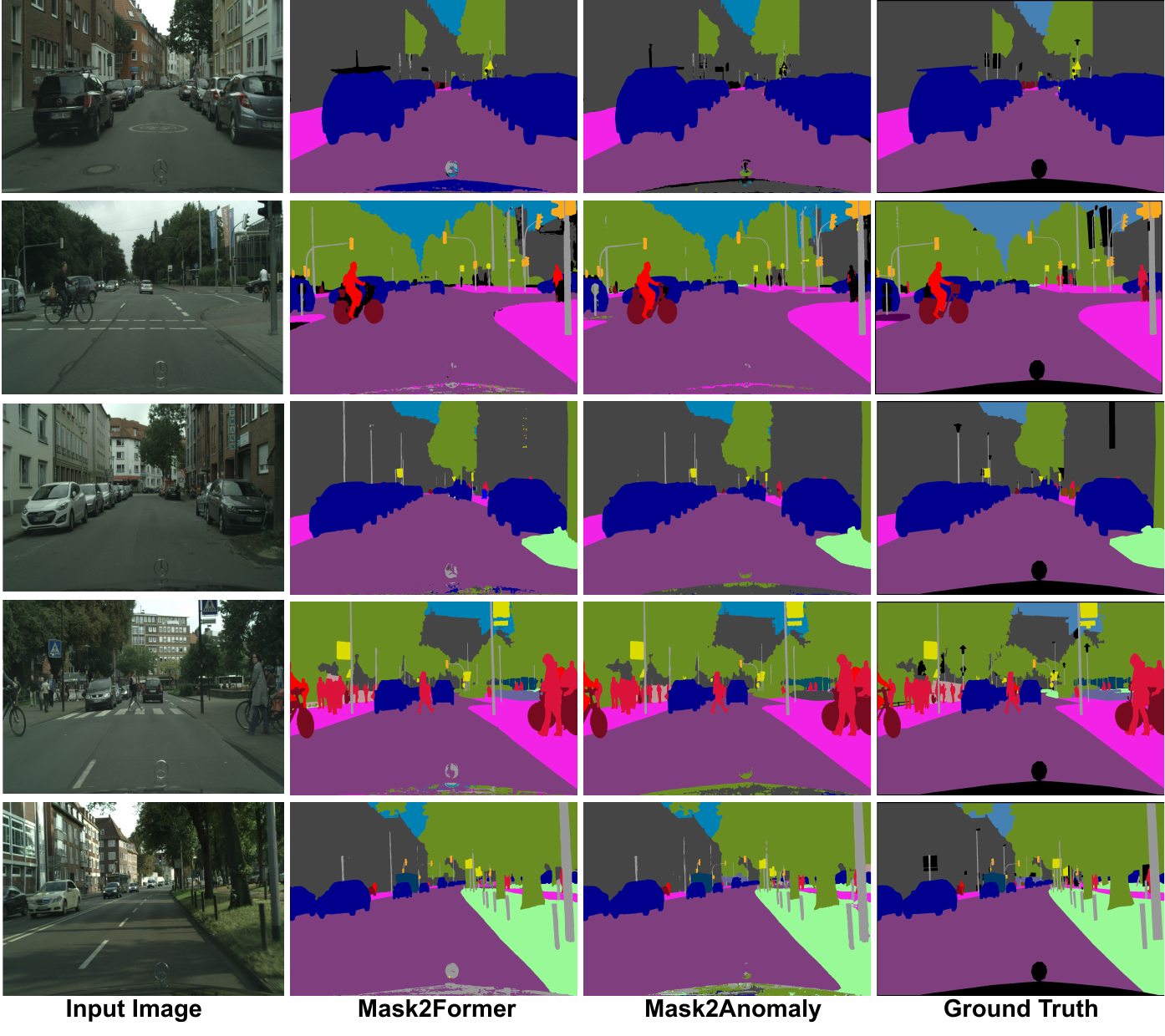}
    \end{center}
    \vspace{-1.5em}
\caption{\textbf{Semantic segmentation results:} We can visually infer that Mask2Anomaly shows similar segmentation results when compared with Mask2Former~\cite{cheng2022masked}.}
    \label{fig:semseg}
    \vspace{-1em}
\end{figure}
\myparagraph{Open-set semantic segmentation:} We the use Streethazard~\cite{hendrycks2019scaling} dataset to train~\our{} with a Swin-Base backbone. The model was trained for 50 thousand iterations keeping all the parameters the same as anomaly segmentation. Next, we train~\our{} on outlier images in a contrastive settings. The outlier image was created by AnomalyMix~\cite{tian2022pixel} using MS-COCO~\cite{lin2014microsoft} and Streethazard image. We train the network for 5000 iterations keeping the Swin-Base backbone frozen. The image crop size was kept at $380\times760$, the rest all the other hyper-parameters were the same as for anomaly segmentation.

\myparagraph{Open-set panoptic segmentation:} We train~\our{} having ResNet-50 backbone for 370 thousand iterations. Our training approach employs a batch size of 8, incorporating cropped input images sized at 640$\times$640. We keep the remaining hyperparameters same as specified in the anomaly segmentation. Across all the three training datasets, which contain 5\%, 10\%, and 15\% of unknown classes, the number of connected components were 2, 2, and 3 respectively. The number of iteration for connected component algorithm was kept at 500 for each training dataset.

\begin{table*}[!ht]
\centering
\renewcommand{\arraystretch}{1.15}
\resizebox{1\linewidth}{!}{\begin{tabular}{c|ccc|ccc|ccc|ccc}
\multicolumn{1}{c|}{}&\multicolumn{3}{c|}{SMIYC RA-21}&\multicolumn{3}{c|}{SMIYC RO-21} &\multicolumn{3}{c|}{Lost \& Found} &\multicolumn{3}{c|}{Average}\\
\cline{1-4}\cline{5-7}\cline{8-10}\cline{11-13}
\multicolumn{1}{c|}{Methods}&\multicolumn{1}{c|}{sIoU $\uparrow$}&\multicolumn{1}{c|}{PPV $\uparrow$} &\multicolumn{1}{c|}{$F1^{*}$$\uparrow$}&{sIoU $\uparrow$}&\multicolumn{1}{c|}{PPV $\uparrow$} &\multicolumn{1}{c|}{$F1^{*}$$\uparrow$}&{sIoU $\uparrow$}&\multicolumn{1}{c|}{PPV $\uparrow$} &\multicolumn{1}{c|}{$F1^{*}$$\uparrow$}&{sIoU $\uparrow$}&\multicolumn{1}{c|}{PPV $\uparrow$} &\multicolumn{1}{c|}{$F1^{*}$$\uparrow$}\\
\Xhline{3\arrayrulewidth}

Max Softmax~\cite{hendrycks2016baseline}(ICLR'17)  &			15.48 &	 	15.29 	 &	5.37 	 &	19.72  &		15.93  &		6.25 &14.20 &62.23 &10.32 &16.47 &31.15 &7.31\\
Ensemble~\cite{lakshminarayanan2017simple}(NurIPS'17)&	16.44 	 &	20.77  &		3.39  &		8.63 	 &	4.71  &		1.28 &6.66 &7.64 &2.68 &10.58 &11.04 &2.45\\
Mahalanobis~\cite{lee2018simple}(NeurIPS'18)   &		14.82  &		10.22 &	 	2.68  &		13.52 	 &	21.79 	 &	4.70  &33.83 &31.71 &22.09 &20.72 &21.24 &9.82\\
Image Resynthesis~\cite{lis2019detecting}(ICCV'19) &	39.68	 &	10.95	 &	12.51	 &	16.61	 &	20.48 &		8.38  &27.16 &30.69 &19.17 &27.82 &20.71 &13.35\\
MC Dropout~\cite{mukhoti2018evaluating}(CVPR'20)   &	20.49 	 &	17.26 	 &	4.26 	 &	5.49 	 &	5.77 	 &	1.05  &17.35 &34.71 &12.99 &14.44 &19.25 &6.10\\
Learning Embedding~\cite{blum2021fishyscapes}(IJCV'21)    &		33.86	 &	20.54 	 &	7.90  &			35.64  &		2.87  &		2.31  &27.16 &30.69 &19.17 &32.22 &18.03 &9.79\\
SML~\cite{jung2021standardized}(ICCV'21)    &	26.00 &	 	24.70  &		12.20 	 &	5.10  &		13.30  &		3.00 & 32.14 &27.57 &26.93 &21.08 &21.86 &14.04\\
SynBoost~\cite{di2021pixel}(CVPR'21)  &		34.68	 &	17.81	 &	9.99	 &	44.28	 &	41.75	 &	37.57 &36.83 &\textbf{72.32} &48.72 &38.60 &43.96 &32.09\\
Maximized Entropy~\cite{chan2021entropy}(ICCV'21) &		49.21&	\underline{39.51}	&28.72	&\underline{47.87}	&\underline{62.64}	&48.51 &45.90 &63.06 &49.92 &47.66 &\underline{55.07} &42.38
\\
JSRNet~\cite{vojir2021road}(ICCV'21) &		20.20  &		29.27 	 &	13.66 	 &	18.55 	 &	24.46  &		11.02 	&34.28 &45.89 &35.97 &24.34 &33.21 &20.22
\\
Void Classifier~\cite{blum2021fishyscapes}(IJCV'21)&	21.14 	 &	22.13  &		6.49  &		6.34  &		20.27  &		5.41  &1.76 &35.08 &1.87 &9.75 &25.83 &4.59
\\
Dense Hybrid~\cite{grcic2022densehybrid}(ECCV'22)&	\underline{54.17} &		24.13 &		\underline{31.08} &		45.74	 &	50.10	 &	\underline{50.72}	&\underline{46.90} &52.14 &\underline{52.33} &\underline{48.94} &42.12 &\underline{44.71}
\\
PEBEL~\cite{tian2022pixel}(ECCV'22) &	38.88	 &	27.20	 &	14.48	 &	29.91	 &	 7.55 &		5.54 &33.47 &35.92 &27.11	&34.09 &23.56 &15.71
\\
\hline
Mask2Former~\cite{cheng2022masked}    &		25.20 	 &	18.20  &		15.30  &		5.00 	 &	21.90  &		4.80 &17.88 &18.09 &9.77  &16.03 &19.40 &9.96
\\
\textbf{Mask2Anomaly (Ours)}   &		\textbf{60.40}  &		\textbf{45.70}  &		\textbf{48.60}  &	 	\textbf{61.40}  &		\textbf{70.30}  &		\textbf{69.80} &	 	\textbf{56.07}  &		\underline{63.41}  &		\textbf{62.78} 	  &\textbf{59.29} &\textbf{59.80} &\textbf{60.39}
\\
\end{tabular}}
\caption{\textbf{Anomaly segmentation component level evaluation:} Mask2Anomaly achieves large improvement on component level evaluation metrics among the baselined methods. Higher values of sIoU, PPV, and $F1^{*}$  are better. The best and second best results are \textbf{bold} and \underline{underlined}, respectively.
} 
\vspace{-1em}
\label{tab:comp-eval} 
\end{table*}

\begin{table*}[!ht]
\centering
\renewcommand{\arraystretch}{1.15}
\resizebox{0.85\linewidth}{!}{\begin{tabular}{c|cc|c|ccc}
\multicolumn{1}{c|}{}&\multicolumn{2}{c|}{Anomaly Segmentation}&\multicolumn{1}{c|}{Close Set Performance}&\multicolumn{3}{c|}{Open Set Performance}\\
\cline{1-4}\cline{5-7}
\multicolumn{1}{c|}{Methods}&\multicolumn{1}{c|}{AuPRC $\uparrow$}&\multicolumn{1}{c|}{FPR$_{95}$ $\downarrow$} &\multicolumn{1}{c|}{mIoU$\uparrow$}&{Open-IoU$^{t5}$ $\uparrow$}&\multicolumn{1}{c|}{Open-IoU$^{t6}$ $\uparrow$} &\multicolumn{1}{c|}{Open-IoU$\uparrow$}\\
\Xhline{3\arrayrulewidth}

MSP~\cite{hendrycks2016baseline} (ICLR'17)  & 7.5 &	27.9 	 &	65.0	 &	32.7  & 40.2  & 35.1 \\
ODIN~\cite{liang2017enhancing} (ICLR'18)  & 7.0&	28.7 	 &	65.0	 &	26.4  & 33.9  & 28.8 \\
Outlier Exposure~\cite{hendrycks2018deep} (ICLR'19) & 14.6 &	17.7 	 &	61.7	 &	43.7  & 44.1  & 43.8 \\
OOD-Head~\cite{bevandic2019simultaneous} (GCPR'19)  & 19.7 &	56.2 	 & \underline{66.6}	&33.7 &34.3  &33.9 \\

MC Dropout~\cite{mukhoti2018evaluating} (CVPR'20)  &	7.5 &	79.4 	 &	-	 &	-  & -  & - \\
SynthCP~\cite{xia2020synthesize} (ECCV'20)  &	9.3 &	28.4 	 &	-	 &	-  & -  & - \\
TRADI~\cite{franchi2020tradi} (ECCV'20) & 7.2 &	25.3 	 &	-	 &	-  & -  & - \\
OVNNI~\cite{franchi2020one} (CoRR'20) & 12.6 &	22.2 	 &	54.6	 &	-  & -  & - \\
Energy~\cite{liu2020energy} (NurIPS'20)  & 12.9 &	18.2 	 &	63.3	 &	41.7  & 44.9  & 42.7 \\
PAnS~\cite{fontanel2021detecting} (CVPRW'21)  & 8.8 &	23.2 	 &	-	 &	-  & -  & - \\
SO+H~\cite{grcic2020dense} (VISIGRAPP'21)  & 12.7 &	22.2 	 &	59.7	 &	-  & -  & - \\
DML~\cite{cen2021deep} (ICCV'21)  & 14.7 &	17.3 	 &	-	 &	-  & -  & - \\
ReAct~\cite{sun2021react} (NurIPS'21) & 10.9 &	21.2 	 &	62.7	 &	33.0  & 36.2  & 34.0 \\
OH*MSP~\cite{bevandic2021dense} (CoRR'21)  & 18.8 &	30.9 	 & \underline{66.6}	&43.3 &44.2  &43.6 \\

ML~\cite{hendrycks2019scaling} (ICML'22)  & 11.6 &	22.5 	 &	65.0	 &	39.6  & 44.5  & 41.2 \\
DenseHybrid~\cite{grcic2022densehybrid} (ECCV'22)  & \underline{30.2} & \textbf{13.0} 	 &	63.0	&\underline{46.1} &\underline{45.3}  &\underline{45.8} \\
\hline
Mask2Anomaly  & \textbf{58.1} & \underline{14.9} 	 &	\textbf{72.3}	&\textbf{59.9} &\textbf{59.7}  &\textbf{59.8} \\

\end{tabular}}
\caption{\textbf{Open-set semantic segmentation quantitative evaluation:} We observe that~\our{} achieves the best performance on open and close set semantic segmentation matrices. `-' indicates the unavailability of benchmarked results.}
\vspace{-1em}
\label{tab:open-set} 
\end{table*}
\begin{figure}[t]
    \begin{center}
    \rotatebox{90}{\tiny{\hspace{2em}Ground Truth\hspace{5em}Mask2Anomaly\hspace{2.25em}Maximized Entropy~\cite{chan2021entropy}\hspace{1.25em}Dense Hybrid~\cite{grcic2022densehybrid}\hspace{3em}Input Image}}
     \includegraphics[width=0.4\textwidth, height=0.5\textwidth]{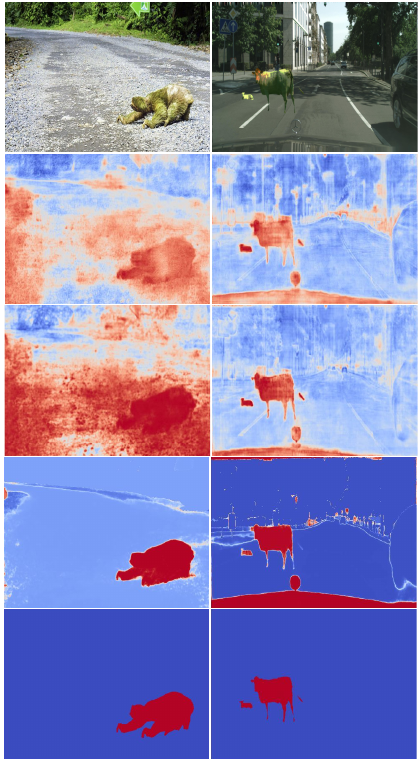}
    \end{center}
    \vspace{-1.25em}
\caption{\textbf{Anomaly segmentation qualitative results}: We observe that per-pixel classification architectures: Dense Hybrid~\cite{grcic2022densehybrid} and Maximized Entropy~\cite{chan2021entropy} suffer from large false positives, whereas~\our{}, which is a mask-transformer, shows accurate pixel-wise anomaly segmentation results. }
    \label{fig:main} \vspace{-1em}
\end{figure}
\begin{figure*}[t]
    \begin{center}
     \includegraphics[width=1\textwidth, height=0.75\textwidth]{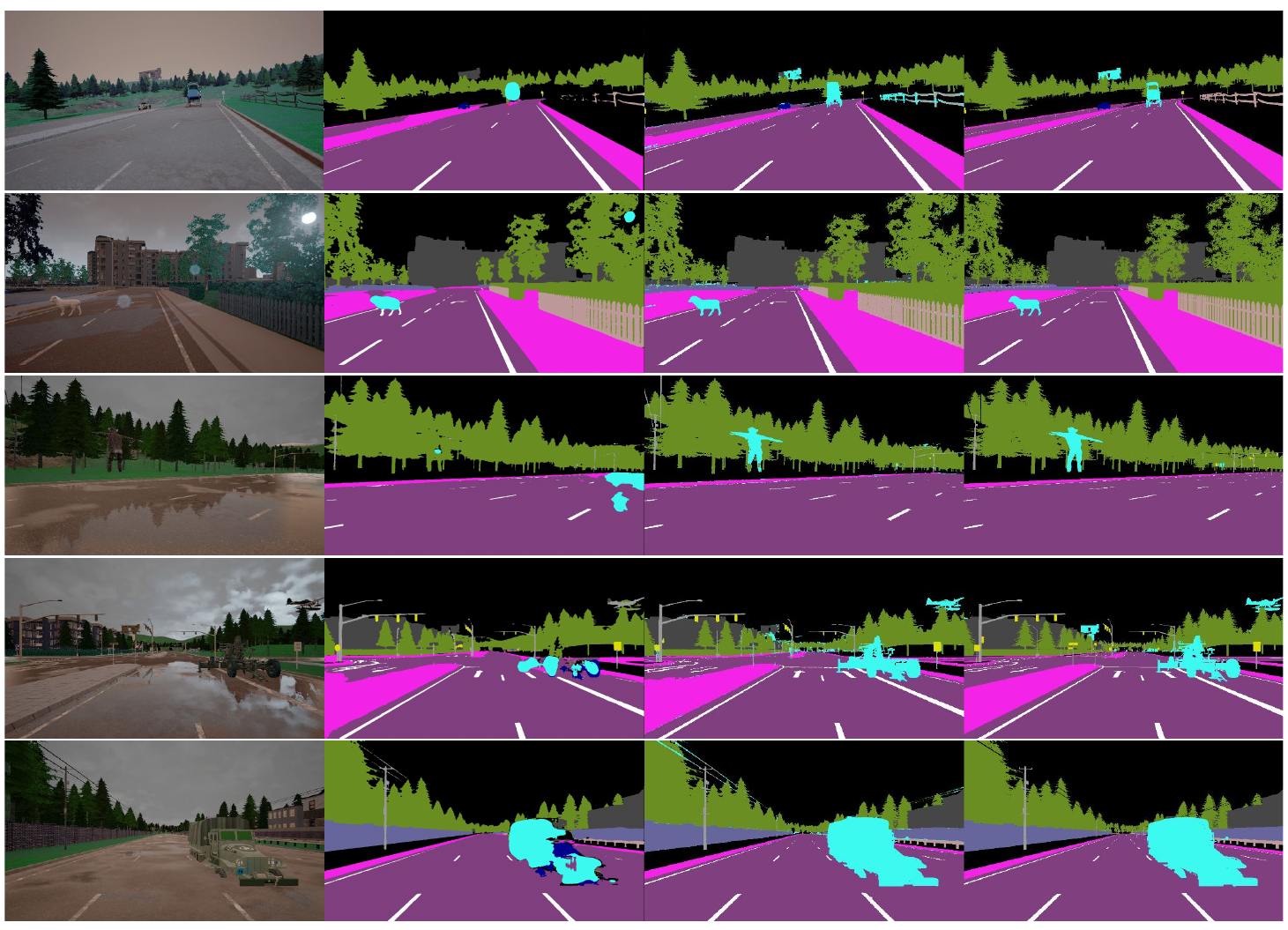}
     \vspace{-1em}
     \rotatebox{0}{{Input Image \hspace{7em}Dense Hybrid~\cite{grcic2022densehybrid}\hspace{6.5em}Mask2Anomaly \hspace{6em}Ground Truth} }
    \end{center}
\caption{\textbf{Qualitative results of open-set semantic segmentation}: We can observe that the~\our{} gives precise boundaries for open-set objects compared to best performing per-pixel architecture i.e. Dense Hybrid.}
    \label{fig:main-open-set} \vspace{-1em}
\end{figure*}
\begin{figure*}[t]
    \begin{center}
     \includegraphics[width=1\textwidth, height=0.45\textwidth]{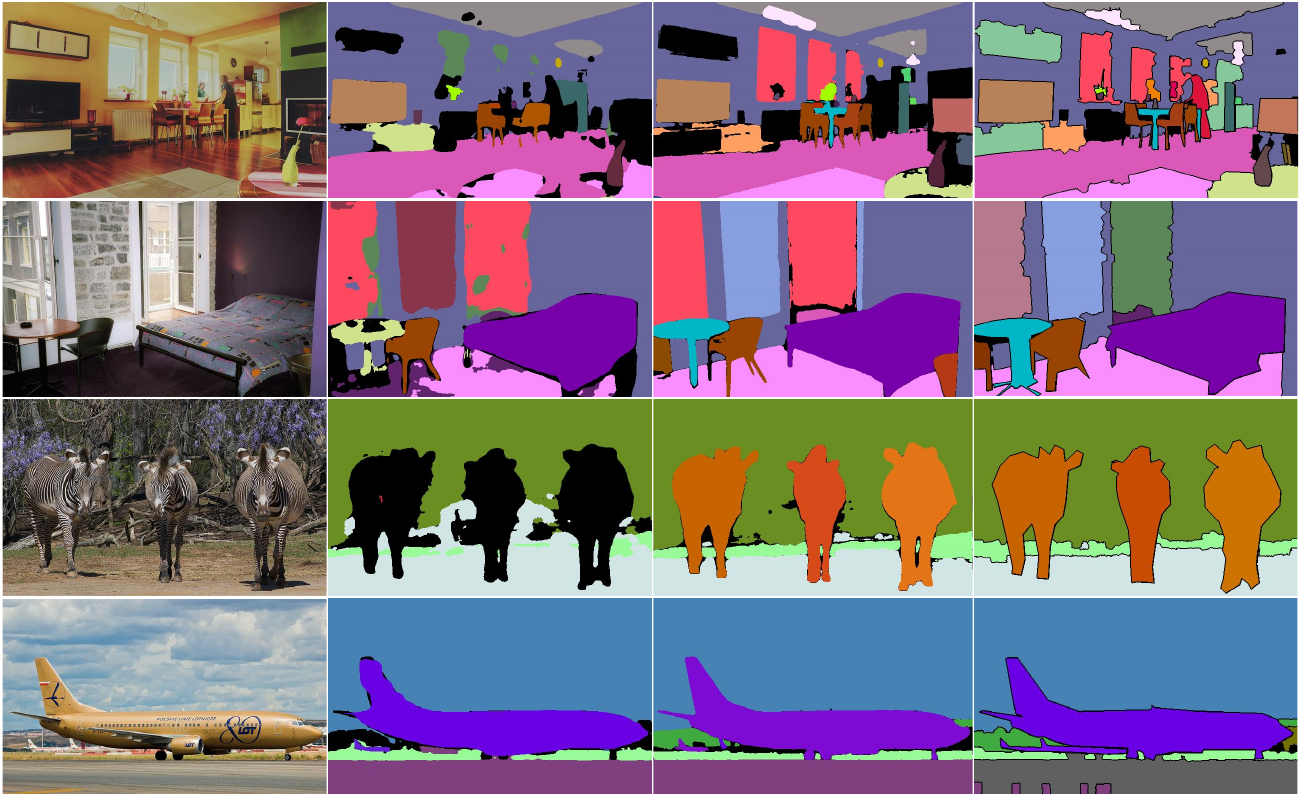}
     \vspace{-1em}
     \rotatebox{0}{{Input Image \hspace{8em}EOPSN~\cite{hwang2021exemplar}\hspace{9em}Mask2Anomaly \hspace{7em}Ground Truth} }
    \end{center}
\caption{\textbf{Open-set panoptic segmentation qualitative results}: Row 1-3: We can observe that~\our{} is better able to segment the different instances of unknown objects compared with the baselined method. Row 4: Shows that~\our{} gives better panoptic segmentation with precise boundaries on known classes.}
    \label{fig:ops} \vspace{-1em}
\end{figure*}
\begin{figure}[t]
\begin{center}
\includegraphics[width=0.9\linewidth, height=0.2\linewidth]{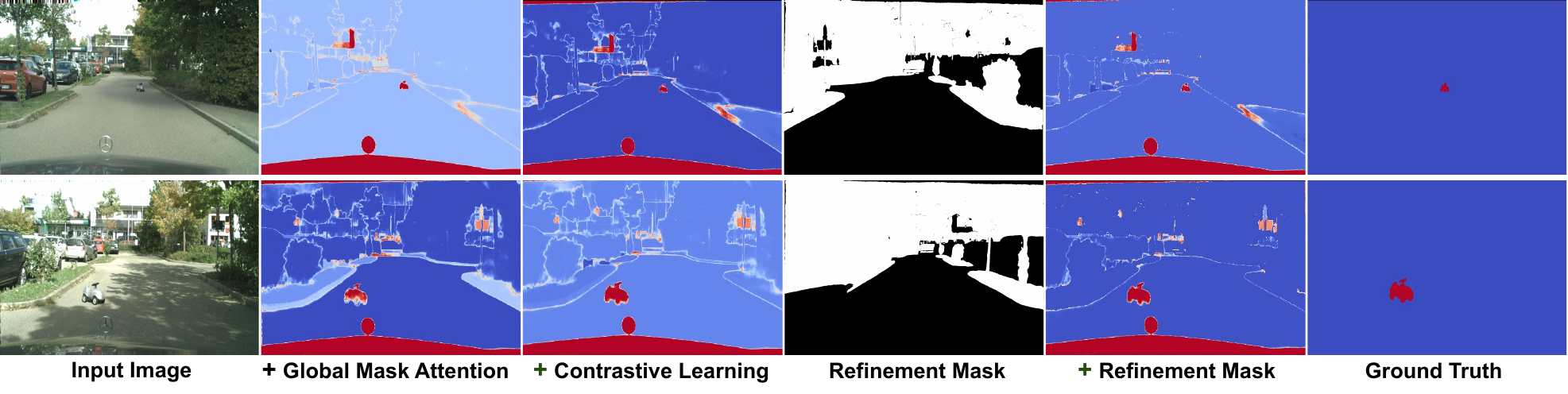}
\end{center}
\vspace{-1.5em}
  \caption{\textbf{Mask2Anomaly Qualitative Ablation}: 
demonstrates the performance gain by progressively adding (left to right ) proposed components. Masked-out regions by refinement mask are shown in white. Anomalies are represented in {red}.}
\label{fig:ablation} \vspace{-1em}
\end{figure}
\begin{table*}[!ht]
\centering
\renewcommand{\arraystretch}{1.15}
\resizebox{1\linewidth}{!}{\begin{tabular}{cc|ccc|ccc|ccc|ccc}
\multicolumn{1}{c|}{}&\multicolumn{1}{c|}{}&\multicolumn{9}{c|}{Known Classes} &\multicolumn{3}{c|}{Unknown Classes}\\
\cline{2-4}\cline{5-7}\cline{8-10}\cline{11-14}
\multicolumn{1}{c|}{K(\%)}&\multicolumn{1}{c|}{Methods}&\multicolumn{1}{c|}{PQ$_{Th}$ $\uparrow$}&\multicolumn{1}{c|}{SQ$_{Th}$ $\uparrow$} &\multicolumn{1}{c|}{RQ$_{Th}$$\uparrow$}&{PQ$_{St}$ $\uparrow$}&\multicolumn{1}{c|}{SQ$_{St}$  $\uparrow$} &\multicolumn{1}{c|}{RQ$_{St}$$\uparrow$}&{PQ $\uparrow$}&\multicolumn{1}{c|}{SQ $\uparrow$} &\multicolumn{1}{c|}{RQ $\uparrow$}&{PQ$_{unk}$ $\uparrow$}&\multicolumn{1}{c|}{SQ$_{unk}$ $\uparrow$} &\multicolumn{1}{c|}{RQ$_{unk}$$\uparrow$}\\
\Xhline{3\arrayrulewidth}

5 &Void-train &43.6	&80.4	&52.8	&28.2 &71.5 &36.0	&37.3 &76.7 &45.9	&8.6 &72.7 &11.8 \\
&EOPSN~\cite{hwang2021exemplar} &44.8 &80.5 &54.2 &28.3 &71.9 &36.2 &38.0 &76.9 &46.8 &23.1 &74.7 &30.9\\
&~\our{}&\textbf{53.0} &\textbf{82.8} &\textbf{63.4} &\textbf{41.3} &\textbf{80.8} &\textbf{50.1} &\textbf{48.2} &\textbf{81.9} &\textbf{57.9} &\textbf{24.3} &\textbf{78.2} &\textbf{32.1}\\
\hline
10 &Void-train &43.7 &80.1 &53.1 &28.1 &73.0 &35.9 &37.1 &77.1 &45.8 &8.1 &72.6 &11.2 \\
&EOPSN~\cite{hwang2021exemplar} &44.5 &80.6 &53.8 &28.4 &71.8 &36.2 &37.7 &76.8 &46.3 &17.9 &76.8 &23.3\\
&~\our{} &\textbf{51.8} &\textbf{82.7} &\textbf{62.0} &\textbf{41.6} &\textbf{80.5} &\textbf{50.5} &\textbf{47.4} &\textbf{81.8} &\textbf{57.1} &\textbf{19.7} &\textbf{77.0} &\textbf{25.7}\\
\hline
20 &Void-train &44.1 &80.1 &53.5 &27.9 &71.6 &35.6 &36.8 &76.3 &45.4 &7.5 &72.9 &10.3 \\
&EOPSN~\cite{hwang2021exemplar} &45.0 &80.3 &54.5 &28.2 &71.2 &36.2 &37.4 &76.2 &46.2 &11.3 &73.8 &15.3\\
&~\our{} &\textbf{50.8} &\textbf{81.6} &\textbf{60.8} &\textbf{40.4} &\textbf{80.8} &\textbf{49.0} &\textbf{46.1} &\textbf{81.2} &\textbf{55.5} &\textbf{14.6} &\textbf{76.2} &\textbf{19.1}\\
\end{tabular}}
\caption{\textbf{Open-set panoptic segmentation quantitative results:} We show quantitative results on the COCO val set by all the methods on varying known-unknown splits. K denoted the \% of unknown classes present in the dataset. Best results are highlighted in bold.} 
\vspace{-1em}
\label{tab:OPS} 
\end{table*}

\subsection{Main Results}
\myparagraph{Anomaly Segmentation:} \Cref{tab:main} shows the pixel-level anomaly segmentation results achieved by {\our} and recent SOTA methods on Fishyscapes, SMIYC, and Road Anomaly datasets. We can observe that~\our{} significantly improves the average AuPRC by 20\% and the FPR$_{95}$ by 60\% compared to the second-best method. 
Another observation is that anomaly segmentation methods based on per-pixel architecture, such as JSRNet, perform exceptionally well on the Road Anomaly dataset. However, JSRNet does not generalize well on other datasets. On the other hand,~\our{} yields excellent results on all the datasets.
Moreover, the property of our mask architecture to encourage objectness rather than individual pixel anomalies, not only reduces the false positive but also improves the localization of whole anomalies. Indeed, \cref{tab:comp-eval} demonstrates that \our{}  outperforms all the baselined methods on component-level evaluation metrics. To conclude, {\our} yields state-of-the-art anomaly segmentation performance both in pixel and component metrics. To get a better understanding of the visual results, in \cref{fig:main} we visually compare the anomaly scores predicted by {\our} and its closest competitors: Dense Hybrid~\cite{grcic2022densehybrid} and Maximized Entropy~\cite{chan2021entropy}. 
The results from both: Dense Hybrid and Maximized Entropy exhibit a strong presence of false positives across the scene, particularly on the boundaries of objects (``things'') and regions (``stuff''). On the other hand, \our{} demonstrates the precise segmentation of anomalies while at the same time having minimal false positives.

Another critical characteristic of any anomaly segmentation method is that it should not disturb the in-distribution classification performance, or else it would make the semantic segmentation model unusable. We show that~\cref{tab:ab}(c)~\our{} achieves mIoU of 80.45, consisting of only GMA as a novel component. However, after mask contrastive training, we find that {\our} maintains an in-distribution accuracy of 78.88 mIoU on the Cityscapes validation dataset, which is still 1.46 points higher than the vanilla Mask2Former. Moreover, it is important to note that both~\our{} and Mask2Former are trained for 90k iterations, indicating that, although ~\our{} additionally attends to the background mask region, it shows convergence similar to Mask2Former.~\cref{fig:semseg} qualitatively shows~\our{} semantic segmentation results are almost identical to Mask2Former.\\

\myparagraph{Open-set semantic segmentation}:~\Cref{tab:open-set} illustrates the open-set semantic segmentation performance of~\our{} on the StreetHazards test set. In terms of anomaly segmentation performance, we observe that~\our{} gives a significant gain of~90\% compared to DenseHybrid in AuPRC with minimal increase in false positives. Notably,~\our{} also gives the best closed set performance, indicating its ability to improve in-distribution while giving state-of-the-art anomaly segmentation results. Furthermore, we measure open-set semantic segmentation using Open-IoU metrics, which allows us to measure anomalous and in-distribution class performance jointly. 

The Streethazard test dataset consists of two sets: t5 and t6. So, to calculate Open-IoU on t5: Open-IoU$^{t5}$, we select the anomaly threshold from t6 at a true positive rate of 95\% and
then re-calculate the classification scores of in-distribution classes t5. We repeat the same steps to get Open-IoU$^{t6}$. To get the overall Open-IoU on the Streethazard test, we calculate the weighted average of Open-IoU on t5 and t6 according to the number of images in each set. In~\Cref{tab:open-set}, we can observe~\our{} outperforms other baselined methods by a significant margin of 30\% on Open-IoU metrics. It is also important to note that methods such as OOD-Head achieve good close-set performance but show low Open-IoU. On the other hand, Outlier Exposure has a relatively better Open-IoU but losses close set performance.~\our{} does not suffer such shortcomings and gives the best open and closed set performances. Qualitatively from~\cref{fig:main-open-set}, we can visually infer that~\our{} is able to preciously segment the anomalous/open-set objects as compared to the best per-pixel architecture i.e., Dense Hybrid.

\myparagraph{Open-set panoptic segmentation}:~\Cref{tab:OPS} summarises the open-set panoptic segmentation performance of all the methods. Void-train is a baseline method in which we train the void regions of an image by treating it as a new class. We can observe~\our{} shows the best open-set panoptic segmentation results among all the baselined methods on different proportions of unknown classes. Additionally, it also shows strong results on in-distribution classes that are indicated by various panoptic evaluation metrics.~\Cref{fig:ops} illustrates the qualitative comparison of~\our{} with baselined methods on most challenging dataset having 20 \% unknown classes. In ~\Cref{fig:ops} (Row: 1-3), we can see~\our{} can better perform panoptic segmentation on unknown instances compared with baselined methods.~\Cref{fig:ops} (Row: 4), shows the panoptic segmentation on known classes where we can observe~\our{} outputs are precise with minimal false positives. 

\subsection{Ablations}
\label{sec:ablation}
All the results reported in this section are based on the FS L\&F validation dataset.

\myparagraph{Mask2Anomaly:} \Cref{tab:ab}(a) presents the results of a component-wise ablation of the technical novelties included in {\our}. We use Mask2Former as the baseline. As shown in the table, removing any individual component from {\our} drastically reduces the results, thus proving that their individual benefits are complimentary. In particular, we observe that the global masked attention has a big impact on the AuPRC and contrastive learning is very important for the FPR$_{95}$. The mask refinement brings further improvements to both. \Cref{fig:ablation} visually demonstrates the positive effect of all the components.
\begin{table*}[t]
\renewcommand{\arraystretch}{1}
      \centering
        \begin{tabular}{ cc }   
        \resizebox{0.3\linewidth}{!}{\label{table-a}\begin{tabular}{ccc|cc}
        GMA & CL & RM & AuPRC$\uparrow$ & FPR$_{95}$$\downarrow$\\
        \Xhline{3\arrayrulewidth}
         &  &   &\textit{10.60} &\textit{89.35}  \\
        \hline
        \cmark  &  &  \cmark &35.05 &87.11  \\
         & \cmark & \cmark  &57.23 &31.93 \\ 
        \cmark  & \cmark &   &68.95 &24.07 \\ 
        \cmark  & \cmark &\cmark  &\textbf{69.41} &\textbf{9.46} \\ 
        \end{tabular}} &  
        \resizebox{0.25\linewidth}{!}{\begin{tabular}{c|cc}
        margin($m$) &  AuPRC$\uparrow$ & FPR$_{95}$$\downarrow$\\
        \Xhline{3\arrayrulewidth}
        $1$   &65.37 &11.61 \\ 
        $0.95$ &65.40 &12.20\\ 
        $0.90$ &66.05 &13.49\\
        $0.80$ &66.20 &14.89\\
        $0.75$ &\textbf{69.41} &\textbf{9.46}\\ 
        $0.50$ &62.07 &13.26\\
        \end{tabular}}\\
        (a) & (b) 
        \\ 
        \end{tabular}

      \centering
        \begin{tabular}{ ccc }   
        \resizebox{0.3\linewidth}{!}{\begin{tabular}{cccc}
        & mIoU$\uparrow$ & AuPRC$\uparrow$ & FPR$_{95}$$\downarrow$\\
        \Xhline{3\arrayrulewidth}
        CA~\cite{cheng2021per}& 76.43  &20.30 &89.35  \\
        MA~\cite{cheng2022masked}& 77.42&10.60 &89.39  \\ 
        \cellcolor{blue!15}{GMA} &\textbf{80.45} &\textbf{32.35} &\textbf{25.95} \\ 
        \end{tabular}} &  
        \resizebox{0.32\linewidth}{!}{\begin{tabular}{c|cc}
        & AuPRC$\uparrow$ & FPR$_{95}$$\downarrow$\\
        \Xhline{3\arrayrulewidth}
        $w/o$ Refinement Mask  &68.95 &24.07 \\
        \hline
        $L_{\{things \hspace{0.1em} \setminus \hspace{0.1em} road\}}$  &67.04 &39.11 \\ 
        
        \cellcolor{blue!15}\textbf{$L_{\{stuff \hspace{0.1em} \setminus \hspace{0.1em} road\}}$}  &\textbf{69.41} &\textbf{9.46} \\ 
        \end{tabular}} &
        
        \resizebox{0.3\linewidth}{!}{\begin{tabular}{c|cc}
        Batch Outlier Probability & AuPRC$\uparrow$& FPR$_{95}$$\downarrow$\\
        \Xhline{3\arrayrulewidth}
        0.1  &63.01 &14.66 \\
        0.2  &\textbf{69.41} &\textbf{9.46}  \\ 
        0.5 &69.20 &11.03 \\ 
        1  &68.77 &10.53 \\ 
        \end{tabular}} \\
        (c) & (d) & (e)\\ 
        \end{tabular}
    \caption{\textbf{\our{} Ablation tables:} \textbf{(a)} Component-wise ablation of Mask2Anomaly. Results in \textit{italics} show Mask2Former results. GMA: Global Mask Attention, CL: Contrastive Learning, and RM: Refinement Mask. \textbf{(b)} Shows the behavior of $L_{CL}$ by choosing different margin($m$) values. We empirically find the best results when $m$ is 0.75.  \textbf{(c)} Global masked attention (GMA) performs the best among various attention mechanisms: Cross-Attention (CA) and Masked-Attention (MA). It is also important to note that the derived results do not have any additional proposed components of~\our{} apart from GMA.  \textbf{(d)} We show the performance gain by using a refinement mask that masks the $ \{stuff \setminus road\} $ regions as anomalies are categorized as $things$ class.  \textbf{(e)} Batch outlier probability is the likelihood of selecting an outlier image for a batch during contrastive training. The best result is achieved at 0.2 probability.    (\textit{All the results reported on FS Lost \& Found validation set}).}
\vspace{-1em}
\label{tab:ab}
\end{table*}

\begin{figure}[t]
\begin{center}
\rotatebox{90}{\tiny{\hspace{3.5em}CA~\cite{cheng2021per}\hspace{3.5em}MA~\cite{cheng2022masked}\hspace{2em}GMA (Ours)}}
\includegraphics[width=0.95\linewidth, height=0.45\linewidth]{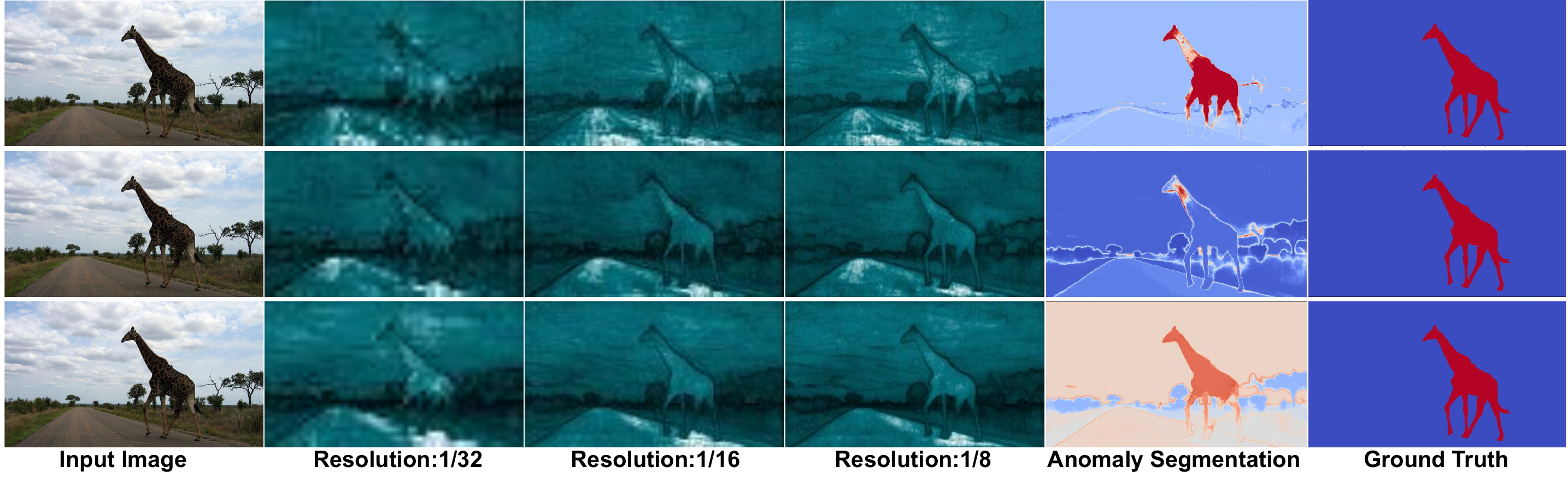}
\end{center} \vspace{-1em}
 \caption{\textbf{Visualization of negative attention maps and results:} Global mask attention gives high attention scores to anomalous regions across all resolutions showing the best anomaly segmentation results among the compared attention mechanisms. Cross-attention performs better than mask-attention but has high false positives and low confidence prediction for the anomalous region. Darker regions represent low attention values. Details to calculate negative attention are given in Section:\ref{sec:ablation}.}
\label{fig:attention}
\end{figure}
\myparagraph{Global Mask Attention:} To better understand the effect of the global masked attention (GMA), in \cref{tab:ab}(c), we compare it to the masked-attention (MA)~\cite{cheng2022masked} and cross-attention (CA)~\cite{vaswani2017attention}. We can observe that although the MA increases the mIoU w.r.t. the CA, it degrades all the metrics for anomaly segmentation, thus confirming our preliminary experiment shown in \cref{fig:mask2former_attention}. On the other hand, the GMA provides improvements across all the metrics.
This is confirmed visually in \cref{fig:attention}, where we show the negative attention maps for the three methods at different resolutions.
The negative attention is calculated by averaging all the queries (since there is no reference known object) and then subtracting one. Note that the GMA has a high response on the anomaly (the giraffe) across all resolutions.

\begin{table}[t]
\small
\centering
\renewcommand{\arraystretch}{1.05}
\resizebox{0.65\linewidth}{!}{\begin{tabular}{c|ccc}
\multirow{1}{*}{Number of Iterations} &\multirow{1}{*}{PQ $\uparrow$} &\multirow{1}{*}{SQ $\uparrow$} &\multirow{1}{*}{RQ $\uparrow$}\\
\hline
100 & 11.4 & 77.2 & 14.8\\
200 & 12.5 & \textbf{77.8} & 16.0\\
500 & \textbf{14.6} & 76.2 & \textbf{19.1}\\
1000 & 10.9 & 76.1 & 14.9\\

\end{tabular}}
\caption{\textbf{Connected component training iteration:} We show the panoptic segmentation performance of unknown classes with the increasing number of iterations. We find the best performance at 500 iterations. Best results are shown in bold.}
\label{tab:itr-ops} \vspace{-1em}
\end{table}
\begin{table}[t]
\small
\centering
\renewcommand{\arraystretch}{1.05}
\resizebox{0.75\linewidth}{!}{\begin{tabular}{c|ccc}
\multirow{1}{*}{Number of Connected Components} &\multirow{1}{*}{PQ $\uparrow$} &\multirow{1}{*}{SQ $\uparrow$} &\multirow{1}{*}{RQ $\uparrow$}\\
\hline
1 & 10.9 & 76.1 & 14.9\\
2 & 12.4 & \textbf{76.5} & 16.3\\
3 & \textbf{14.6} & 76.2 & \textbf{19.1}\\
5 & 14.0 & 78.2 & 17.9\\

\end{tabular}}
\caption{\textbf{Number of connected components:} Shows the panoptic segmentation performance of unknown classes with the increasing number of connected components. We find the best performance at 3. Best results are shown in bold.}
\label{tab:num-cc} \vspace{-1em}
\end{table}
\begin{table}[t]
\small
\centering
\renewcommand{\arraystretch}{1.05}
\resizebox{0.75\linewidth}{!}{\begin{tabular}{c|ccc}
\multirow{1}{*}{} &\multirow{1}{*}{PQ $\uparrow$} &\multirow{1}{*}{SQ $\uparrow$} &\multirow{1}{*}{RQ $\uparrow$}\\
\hline
ours & 14.6 & 76.2 & 19.1\\
\hline
- mining unknown instances & 9.1 & 72.9 & 12.8\\
- global mask attention & 9.0 & 72.3 & 11.3\\

\end{tabular}}
\caption{\textbf{Mining unknown instances for OPS:} We show the negative impact on panoptic segmentation performance of unknown classes by progressively subtracting global mask attention and mining unknown instances components.}
\label{tab:ops-comp} \vspace{-2em}
\end{table}
\begin{table*}[ht]
\centering
\setlength{\tabcolsep}{2pt} 
\renewcommand{\arraystretch}{1.15}

\resizebox{0.8\linewidth}{!}{\begin{tabular}{c|cc|cc|cc|cc|cc}
\multicolumn{1}{c|}{}&\multicolumn{2}{c|}{SMIYC-RA21}&\multicolumn{2}{c|}{SMIYC-RO21}&\multicolumn{2}{c|}{FS L\&F}&\multicolumn{2}{c|}{FS Static}& \multicolumn{2}{c}{Average $\sigma$}\\
\cline{2-3}\cline{4-5}\cline{6-7}\cline{8-9}\cline{10-11}
\multicolumn{1}{c|}{Methods}&\multicolumn{1}{c|}{AuPRC $\uparrow$} &\multicolumn{1}{c|}{FPR$_{95}$ $\downarrow$}&\multicolumn{1}{c|}{AuPRC $\uparrow$} &\multicolumn{1}{c|}{FPR$_{95}$ $\downarrow$}&\multicolumn{1}{c|}{AuPRC $\uparrow$} &\multicolumn{1}{c|}{FPR$_{95}$ $\downarrow$}&\multicolumn{1}{c|}{AuPRC $\uparrow$} &\multicolumn{1}{c|}{FPR$_{95}$ $\downarrow$}&\multicolumn{1}{c|}{AuPRC } &\multicolumn{1}{c}{FPR$_{95}$}\\
\Xhline{3\arrayrulewidth}
Mask2Anomaly-S1  &95.48 &2.41 &92.89 &0.15 &69.41 &9.46 &90.54 &1.98 &- &- \\
Mask2Anomaly-S2  &92.03 &3.22 &92.3 &0.27 &69.19 &13.47 &85.63 &5.06 &- &- \\
$\sigma$(Mask2Anomaly)  &$\pm$ 2.44 &$\pm$0.57 &$\pm$0.42 &$\pm$0.08 &$\pm$0.16 &$\pm$2.84 &$\pm$3.47 &$\pm$2.18 &$\pm$1.62	&$\pm$1.41 \\
\hline

Dense Hybrid-S1 &52.99 &38.87 &66.91 &1.91 &56.89 &8.92 &52.58 &6.03 &- &-\\

Dense Hybrid-S2 &60.59 &32.14 &79.64 &1.01 &47.97 &18.35 &54.22 &5.24 &- &-\\

$\sigma$(Dense Hybrid) &$\pm$5.37 &$\pm$4.76 &$\pm$9.00 &$\pm$0.64 &$\pm$6.31 &$\pm$6.67 &$\pm$1.16 &$\pm$0.56 &$\pm$5.46	&$\pm$3.15
\end{tabular}}
\caption{\textbf{Performance stability of Mask2Former:} We can observe that the average performance deviation in dense hybrid is significantly higher than Mask2Anomaly. $\sigma$ denotes the standard deviation. } \vspace{-1em}
\label{sup-tab:ood-sets} 
\end{table*}

\myparagraph{Refinement Mask:} \Cref{tab:ab}(d) shows the performance gains due to the refinement mask. We observe that filtering out the $\{ \text{``stuff''} \setminus \text{``road''} \} $ regions of the prediction map improves the FPR$_{95}$ by $14.61$ along with marginal improvement in AuPRC. On the other hand, removing the $\{ \text{``things''} \setminus \text{``road''}\}$ regions degrades the results, confirming our hypothesis that anomalies are likely to belong to the ``things'' category. \Cref{fig:ablation} qualitatively shows the improvement achieved with the refinement mask.

\myparagraph{Mask Contrastive Learning:} 
We tested the effect of the margin in the contrastive loss $L_{CL}$, and we report these results in \cref{tab:ab}(b). We find that the best results are achieved by setting $m$ to 0.75, but the performance is competitive for any value of $m$ in the table. Similarly, we tested the effect of the batch outlier probability, which is the likelihood of selecting an outlier image in a batch. The results shown in \cref{tab:ab}(e) indicate that the best performance is achieved at $0.2$, but the results remain stable for higher values of the batch outlier probability.
\begin{table}[t]
\centering
\renewcommand{\arraystretch}{1.15}
\resizebox{1\linewidth}{!}{\begin{tabular}{c|c|ccc|cc}
\multirow{2}{*}{Method} &\multirow{2}{*}{Backbone} &\multirow{2}{*}{AuPRC$\uparrow$} &\multirow{2}{*}{FPR$_{95}\downarrow$}  & \multirow{2}{*}{FLOPs$\downarrow$} & Training $\downarrow$\\\
& & & & & Parameters \\
\Xhline{3\arrayrulewidth}
Mask2Former~\cite{cheng2022masked}& ResNet-50 &10.60 &89.35 &\textbf{226G} &44M\\

& ResNet-101 &9.11 &45.83 &293G &63M\\

& Swin-T &24.54 &37.98 &232G &42M\\
& Swin-S &30.96 &36.78 &313G &69M\\
\hline
Mask2Anomaly$^\ddag$ & ResNet-50 &\textbf{32.35} &\textbf{25.95} &258G &\textbf{23M}\\
\end{tabular}}
\caption{\textbf{Architectural Efficiency of Mask2Anomaly:} Mask2Anomaly outperforms the best performing Mask2Former architecture with Swin-S as backbone by using almost 30\% trainable parameters. Mask2Anomaly$^\ddag$ only uses global mask attention.}
\label{tab:largebackbone} \vspace{-1em}
\end{table}

\myparagraph{Mining Unknowns Instances:} We quantitatively summarise the impact of mining unknown instances in panoptic segmentation of unknown instances shown in~\cref{tab:ops-comp}. We can clearly observe that removing the mining of unknown instances from~\our{} drastically reduces the performance across all the metrics. Also, the absence of global mask attention further degrades performance.

\myparagraph{Connected Components:}~\cref{tab:itr-ops,tab:num-cc} shows the impact of connected components hyperparameters on open-set panoptic segmentation of unknown classes. In both tables, we train the model on dataset split having 20 \% of unknown classes. In~\cref{tab:itr-ops}, we can observe that~\our{} shows the best performance at 500 iterations. Whereas, in~\cref{tab:num-cc}, we achieve the best performance when the number of connected components is set to 3. 

\myparagraph{Architectural Efficacy of Mask2Anomaly:} We demonstrate the efficacy of {\our} by comparing it to the vanilla Mask2Former but using larger backbones. The results in~\cref{tab:largebackbone} show that despite the disadvantage, Mask2Anomaly with a ResNet-50 still performs better than Mask2Former using large transformer-based backbones like Swin-S. It is also important to note that the number of training parameters for Mask2Anomaly can be reduced to $23M$ as we use a frozen self-supervised pre-trained encoder during the entire training, which is significantly less than all the Mask2Former variations.  
\section{Discussion}
\myparagraph{Performance stability:}
Employing an outlier set to train an anomaly segmentation model presents a challenge because the model's performance can vary significantly across different sets of outliers. Here, we show that~\our{} performs similarly when trained on different outlier sets. We randomly chose two subsets of 300 MS-COCO images (S1, S2) as our outlier dataset for training~\our{} and DenseHybrid.~\Cref{sup-tab:ood-sets} shows the performance of~\our{} and Dense Hybrid trained on S1 and S2 outlier sets, along with the standard deviation($\sigma$) in the performance. We can observe that the variation in performance for the dense hybrid is significantly higher than~\our{}. Specifically, in dense hybrid, the average deviation in AuPRC is greater than 300\%, and the average variation in FPR$_{95}$ is more than 200\% compared to~\our{}. 

\begin{figure}[t]
    \centering
    \includegraphics[width=1\linewidth]{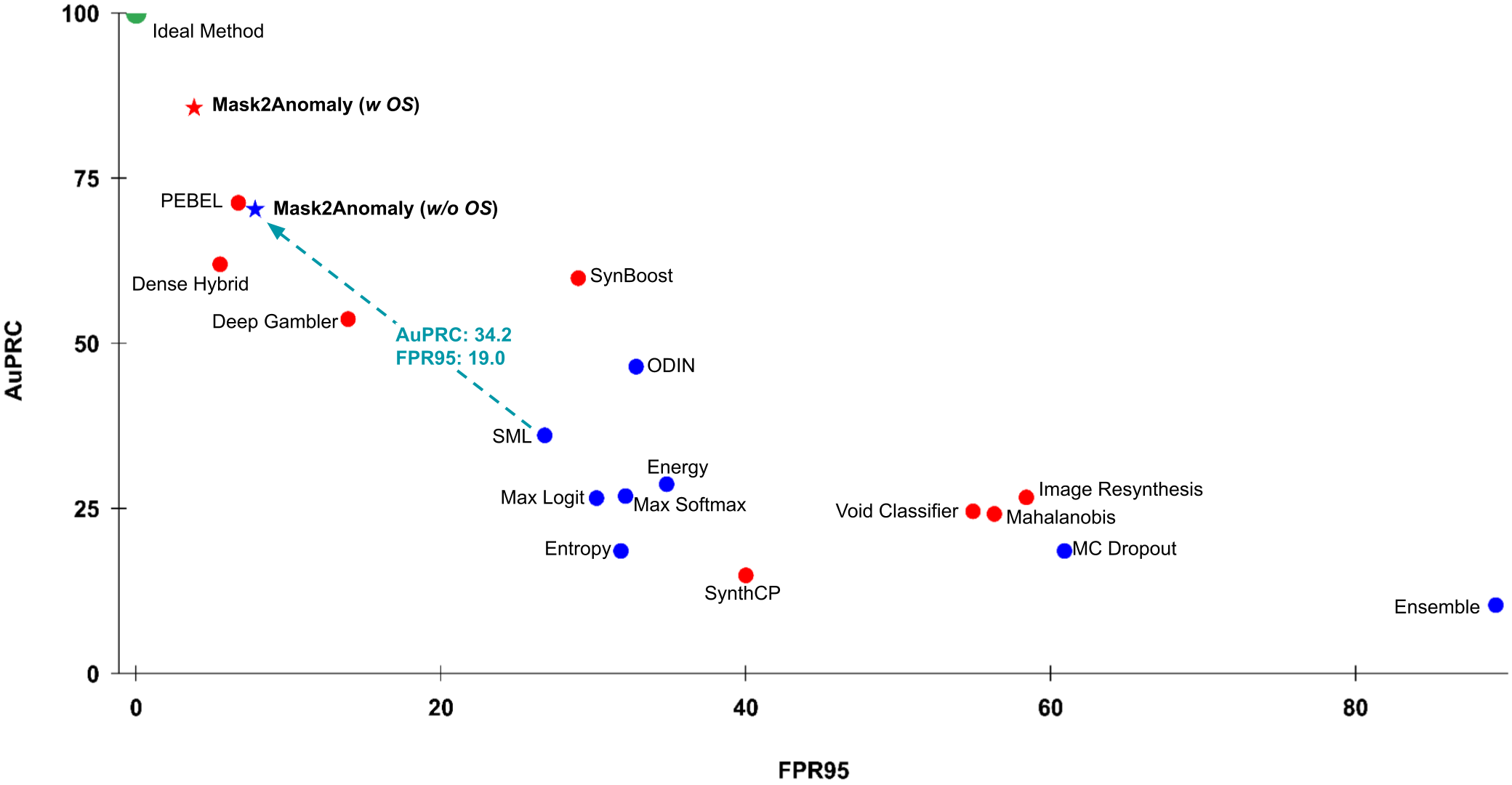}
  \caption{\textbf{Bridging the supervision gap:}
 In this figure, we represent methods that utilize outlier supervision in red, and those without outlier supervision are in blue. We can observe Mask2Anomaly \textit{(w/o OS)}:~\our{} without using outlier supervision, shows significant performance gain among anomaly segmentation methods that do not use any extra supervision. Also, displays a similar performance to PEBEL, which is the best per-pixel method that utilizes additional supervision).}
    \label{fig:gapreduce} \vspace{-1em}
\end{figure}

\myparagraph{Reducing the supervision gap:} In our previous discussion, we show models that are trained with outlier supervision have varying performance across different sets of outliers. So, we extend the previous discussion by demonstrating the performance of~\our{} without reliance on outlier supervision. We evaluate the performance of all the baselined method average over the validation dataset of FS static, FS L\&F, SMIYC-RA21 and SMIYC-RO21. \cref{fig:gapreduce} shows the performance of~\our{} with or without outlier supervision names as ~\our{} \textit{(w OS)} and ~\our{} \textit{(w/o OS)}, respectively. In the plot, we can see unequivocally that {\our} \textit{(w/o OS)} significantly reduces the anomaly segmentation performance gap between the methods with outlier supervision and notably outperforms methods that do not use outlier supervision. 
\begin{figure}[t]
    \begin{center}
        \includegraphics[width=1\linewidth]{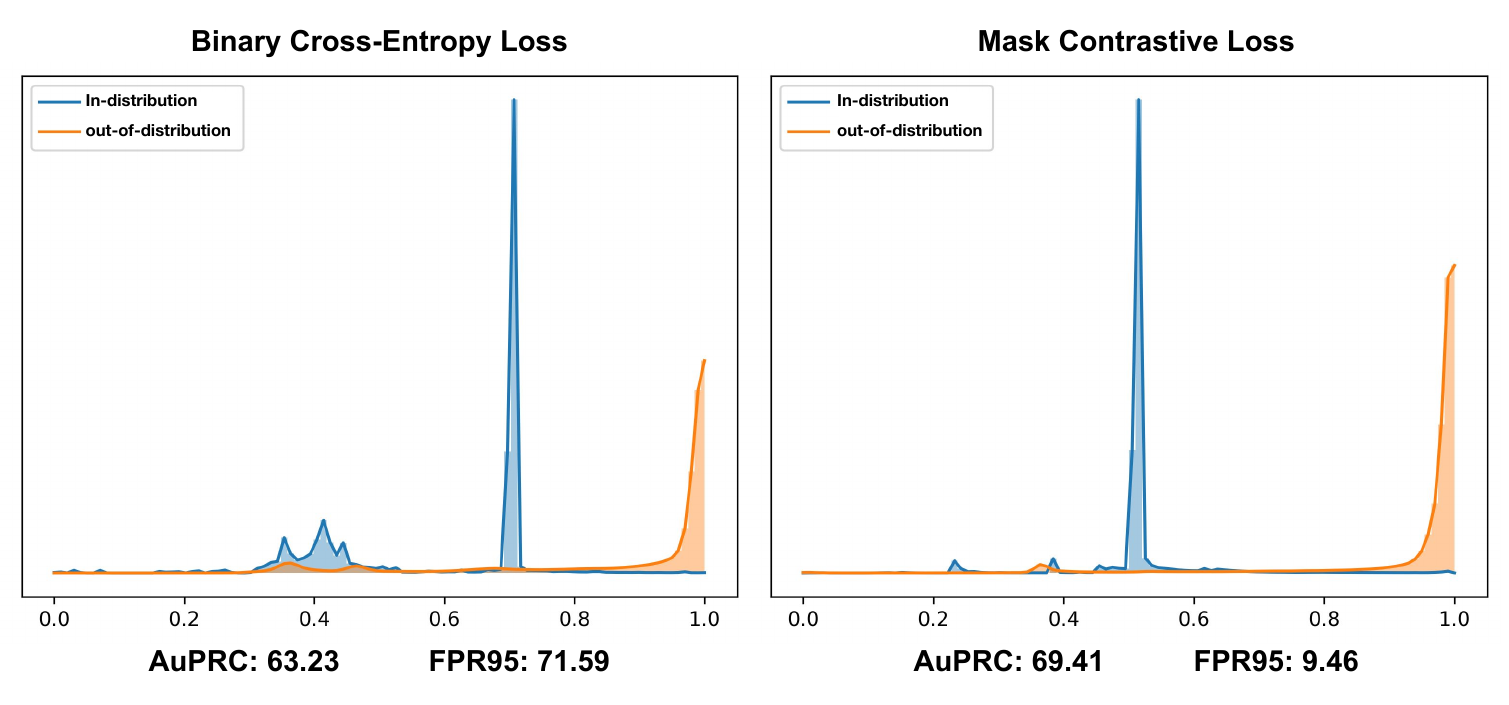}
    \end{center}
    \vspace{-1.5em}
\caption{\textbf{Outlier Loss Comparision:} During the training~\our{}, on the outlier set, we find that incorporating a mask contrastive loss, which is a margin-based loss function, resulted in better performance compared to the conventional binary cross-entropy loss. These experiments were conducted on the FS L\&F validation set.}
    \label{fig:BCEvsCL}
    \vspace{-1em}
\end{figure}

\myparagraph{Outlier Loss:} In this discussion, we will examine the efficacy of mask contrastive loss in anomaly segmentation. We empirically demonstrate why mask contrastive loss, a margin-based loss, performs better at anomaly segmentation by comparing it with binary cross-entropy loss as an outlier loss. So, we train~\our{} with $M_{OOD}$ using binary-cross entropy which equates the outlier loss as:
\begin{equation}
L_{BCE}=M_{OOD}\log(l_{N})+(1-M_{OOD})\log(1-l_{N})
\end{equation}
and, the new total loss at the outlier learning stage becomes:
\begin{equation}
    L_{ood} = L_{BCE} + L_{masks} + \lambda_{ce}L_{ce}
\end{equation}
$l_{N}$ is the negative likelihood of in-distribution classes calculated using the class scores $C$ and class masks $M$.~\Cref{fig:BCEvsCL} illustrates the anomaly segmentation performance comparison on FS L\&F validation dataset between the~\our{} when trained with the binary cross entropy loss and mask contrastive loss, respectively. We can observe that the mask contrastive loss achieves a wider margin between out-of-distribution(anomaly) and in-distribution prediction while maintaining significantly lower false positives.\\

\myparagraph{Global Mask Attention:} The application of global mask attention in semantic segmentation has shown a positive impact on performance, as demonstrated in~\cref{tab:ab}(c). So, we further investigate to assess the generalizability of this positive effect on Ade20K~\cite{zhou2017scene} and Vistas~\cite{neuhold2017mapillary}. To evaluate the possible benefits of global mask attention, we trained the Mask2Former architecture using both masked attention and global masked attention for 40 thousand iterations. Mask2Former performed mIoU scores of 43.20 and 38.17 on masked attention, while global mask attention yields better mIoU scores of 43.80 (+0.6) and 38.92 (+0.75) on Ade20K and Vistas, respectively.
\begin{figure}[t]
    \begin{center}
        \includegraphics[width=1\linewidth]{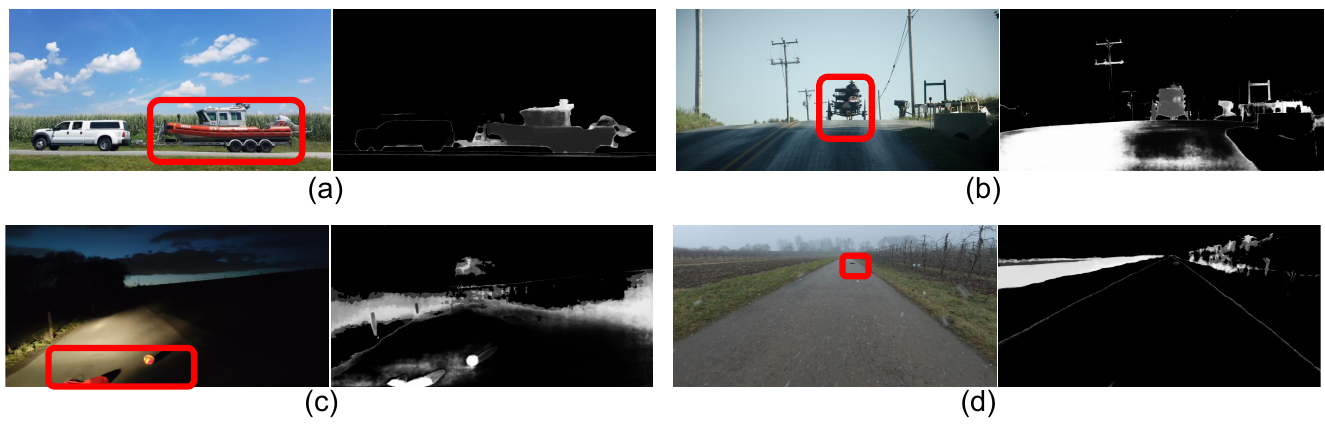}
    \end{center}
    \vspace{-1.5em}
\caption{\textbf{Failure Cases:} {(a, b)~\our{} fail to segment trailer or carriage as they have a similar appearance as car or bus. ~\our{} struggle to perform well in poor illumination (c) and weather conditions (d). The white region represents the anomalies that are enclosed in red boxes.}}
    \label{fig:failurecases}
    \vspace{-1em}
\end{figure}

\myparagraph{Failure Cases:}~\cref{fig:failurecases} illustrates the failure cases predicted by~\our{}. 
 It is apparent that~\our{} faces difficulties when anomalies exhibit a resemblance to in-distribution classes like cars or buses, as shown in in~\cref{fig:failurecases} (a, b). In ~\cref{fig:failurecases} (c) shows increased false positives around anomalies when illumination conditions are poor.  Weather conditions adversely effects~\our{} performance as seen in ~\cref{fig:failurecases} (d). We think that improving anomaly segmentation in such scenarios would be a promising avenue for future research.

\section{Conclusion}
In this work, we introduce Mask2Anomaly, a universal architecture that is designed to jointly address anomaly and open-set segmentation utilizing a mask transformer. Mask2Anomaly incorporates a global mask attention mechanism specifically to improve the attention mechanism for anomaly or open-set segmentation tasks. For the anomaly segmentation task, we propose a mask contrastive learning framework that leverages outlier masks to maximize the distance between anomalies and known classes. Furthermore, we introduce a mask refinement technique aimed at reducing false positives and improving overall performance. For the open-set segmentation task, we developed a
novel approach to mine unknown instances based on mask-architecture properties. Through extensive qualitative and quantitative analysis, we demonstrate the effectiveness of Mask2Anomaly and its components. Our results highlight the promising performance and potential of Mask2Anomaly in the field of anomaly and open-set segmentation. We believe this work will open doors for a new development of novel anomaly and open-set segmentation approaches based on masked architecture, stimulating further advancements in the field.

{\small
\bibliographystyle{ieee_fullname}
\bibliography{egbib}
}

\begin{IEEEbiography}[{\includegraphics[width=1in,height=1.25in,clip,keepaspectratio]{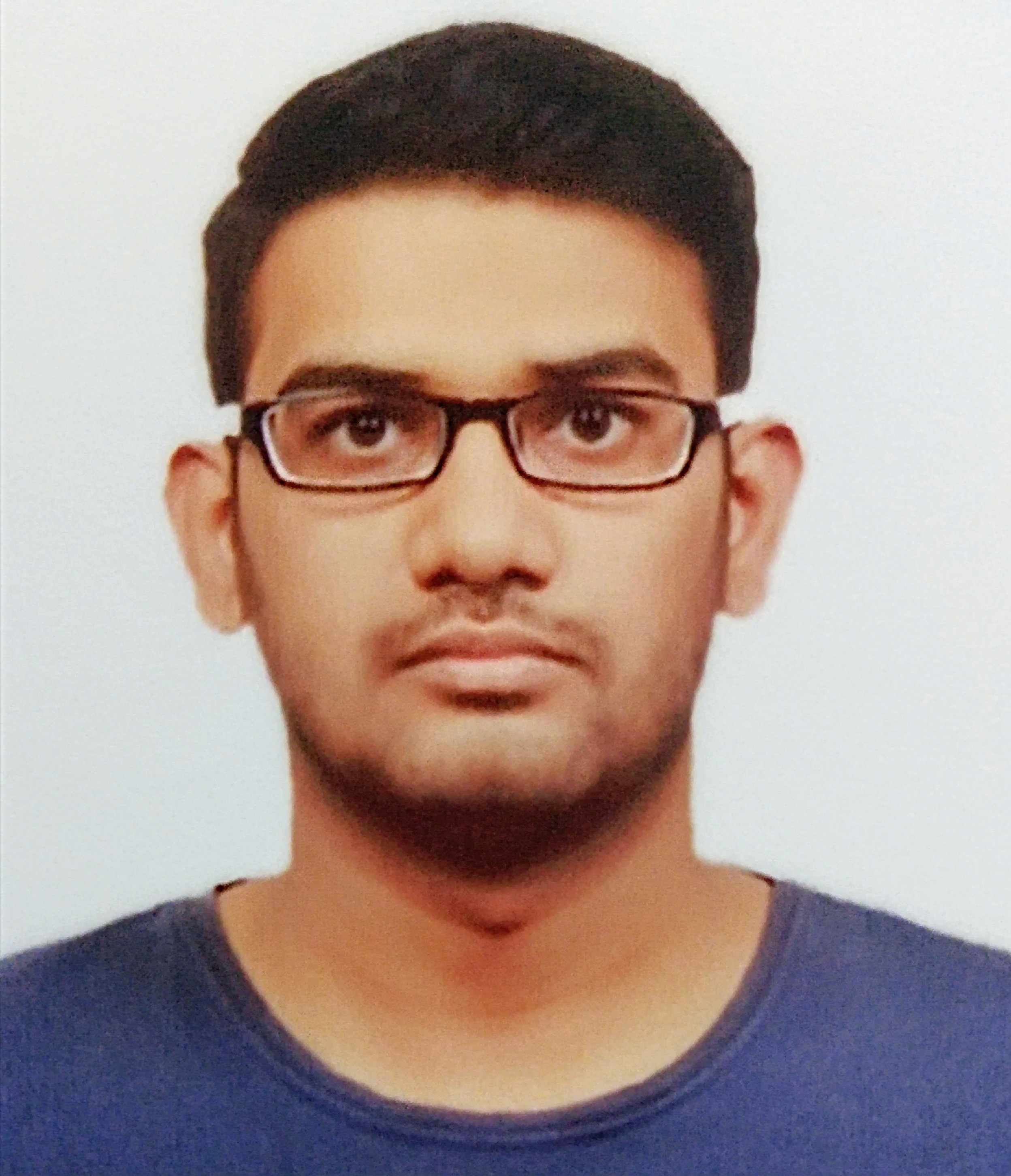}}]{Shyam Nandan Rai} Shyam Nandan Rai is an ELLIS Ph.D. student in Computer and Control Engineering at the Politecnico di Torino, funded by Italian National PhD Program in Artificial Intelligence. He is a member of the Visual Learning and Multimodal Applications Laboratory (VANDAL), supervised by Prof. Barbara Caputo and Prof. Carlo Masone. He received his master's degree in Computer Science at IIIT Hyderabad, in 2020.
\end{IEEEbiography}

\begin{IEEEbiography}[{\includegraphics[width=1in,height=1.25in,clip,keepaspectratio]{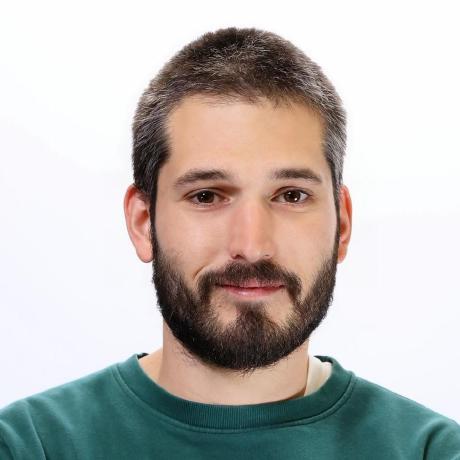}}]{Fabio Cermelli} Fabio Cermelli is a Ph.D. student in Computer and Control Engineering at the Politecnico di Torino, funded by the Italian Insitute of Technology (IIT). He received his master thesis in Software Engineering (Computer Engineering) with honors at the Politecnico di Torino in 2018. He is member of the Visual Learning and Multimodal Applications Laboratory (VANDAL), supervised by Prof. Barbara Caputo. During his first year, he was a visiting Ph.D. student in the Technologies of Vision Laboratory at FBK.\end{IEEEbiography}

\begin{IEEEbiography}[{\includegraphics[width=1in,height=1.25in,clip,keepaspectratio]{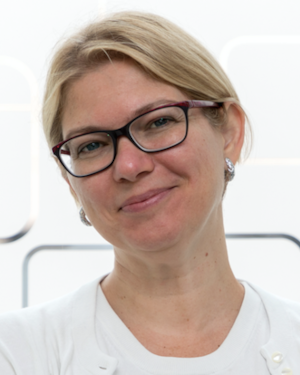}}]{Barbara Caputo}
Barbara Caputo received the Ph.D. degree in computer science from the Royal Institute of Technology (KTH), Stockholm, Sweden, in 2005. From 2007 to 2013, she was a Senior Researcher with Idiap-EPFL (CH).
Then, she moved to Sapienza Rome University thanks to a MUR professorship, and joined the Politecnico di Torino, in 2018. Since 2017, she has been a double affiliation with the Italian Institute of Technology (IIT). She is currently a Full Professor with the Politecnico of Torino, where she leads the Hub AI@PoliTo.
She is one of the 30 experts who contributed to write the Italian Strategy on AI, and coordinator of the Italian National Ph.D. on AI and Industry 4.0, sponsored by MUR.
She is also an ERC Laureate and an ELLIS Fellow.
Since 2019, she serves on the ELLIS Board.
\end{IEEEbiography}

\begin{IEEEbiography}[{\includegraphics[width=1in,height=1.25in,clip,keepaspectratio]{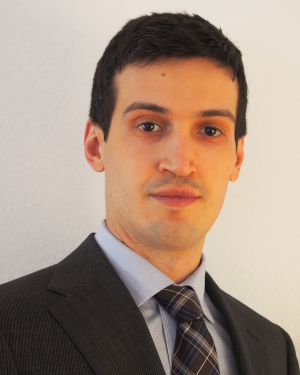}}]{Carlo Masone} (Member, IEEE)
Carlo Masone is currently an Assistant Professor at Politecnico di Torino. He received his B.S. degree and M.S. degree in control engineering from the Sapienza University, Rome, Italy, in 2006 and 2010 respectively, and he received his Ph.D. degree in control engineering from the University of Stuttgart in collaboration with the Max Planck Institute for Biological Cybernetics (MPI-Kyb), Stuttgart, Germany, in 2014. 
From 2014 to 2017 he was a postdoctoral researcher at MPI-kyb, within the Autonomous Robotics \& Human-Machine Systems group. From 2017 to 2020 he worked in industry on the development of self-driving cars. From 2020 to 2022 he was a senior researcher at the Visual and Multimodal Applied Learning, at Politecnico di Torino.
\end{IEEEbiography}

\end{document}